\documentclass[12pt]{article}
\usepackage{amsmath,amssymb,booktabs,array,multirow,graphicx}
\usepackage{textcomp,authblk}
\usepackage{geometry}
\title{An even-load-distribution design for composite bolted joints using a novel circuit model and neural network}

\author[1]{Cheng Qiu}
\author[1]{Yuzi Han}
\author[1]{Logesh Shanmugam}
\author[2]{Fengyang Jiang}
\author[2*]{Zhidong Guan}
\author[3*]{Shanyi Du}
\author[1*]{Jinglei Yang}

\affil[1]{Department of Mechanical and Aerospace Engineering, Hong Kong University of Science and Technology, Hong Kong SAR, China}
\affil[2]{School of Aeronautic Science and Engineering, Beihang University, Beijing, China}
\affil[3]{Center for Composite Materials and Structures, Harbin Institute of Technology, Harbin, China}
\date{}

\begin{document}

\maketitle

\begin{abstract}
Due to the brittle feature of carbon fiber reinforced plastic laminates, mechanical multi-joint within these composite components shows uneven load distribution for each bolt, which weaken the strength advantage of composite laminates. In order to reduce this defect and achieve the goal of even load distribution in mechanical joints, we propose a machine learning-based framework as an optimization method. Since that the friction effect has been proven to be a significant factor in determining bolt load distribution, our framework aims at providing optimal parameters including bolt-hole clearances and tightening torques for a minimum unevenness of bolt load. A novel circuit model is established to generate data samples for the training of artificial networks at a relatively low computational cost. A database for all the possible inputs in the design space is built through the machine learning model. The optimal dataset of clearances and torques provided by the database is validated by both the finite element method, circuit model, and an experimental measurement based on the linear superposition principle, which shows the effectiveness of this general framework for the optimization problem. Then, our machine learning model is further compared and worked in collaboration with commonly used optimization algorithms, which shows a potential of greatly increasing computational efficiency for the inverse design problem.
\end{abstract}

\section{Introduction}
Composite laminates, especially carbon fiber reinforced plastics (CFRP), has been widely used in aerospace industries, due to the excellent strength-to-weight ratio and high stiffness-to-weight ratio. Although integral-forming of composite components may take full advantages of its load-carrying ability, joints among the CFRP parts are inevitable where different connection techniques may be used including mechanically fastened joints, adhesively bonded joints, and hybrid joints\cite{thoppul2009mechanics}. Mechanically fastened joints, which dominates among these techniques in their usage in aerospace structure, hold the advantage of easily-removability, non-sensitivity to the environment, and high load-carrying ability\cite{katnam2013bonded, raju2016improving}. 

Although the design process of metal joints has been well established, it has to be faced with some differences due to the brittle feature of CFRP laminates. While the plasticity of metal material renders it the load redistribution ability, the load distribution ratio is almost unchanged for CFRP mechanical joint before ultimate failure. Generally, the design methodology or strength validation of the mechanical joint takes the following steps\cite{camanho2006design,xiao2005bearing,xiao2005bearing2}. First, the stress conditions of the multi-bolt joint should be extracted from an overall load analysis on global structures. Second, the load distribution would be determined based on the coordination relationship among the deformations of each joined part. After acquiring each bolt load, detailed stress analysis is going to be performed on the area where damage is most likely to occur\cite{puck2002failure,pinho2006physically}. 

Determining bolt load distribution is one of the research hotspots in composite joints since it plays a fundamental role in structural analysis\cite{mccarthy2006simple,lecomte2014analytical,yang2018enhanced}. The typical problem is a joint with single-column, multi-row bolts under external tensile load. Experimental measurement of bolt load is difficult as a result of the narrow space at the contact area for the placement of sensors like strain gauges\cite{liu2018interpretation}. Finite element method or analytical method with engineering experience parameters could be used but faced with drawbacks of computational efficiency and accuracy\cite{lecomte2014analytical, zhao2019modified}. The spring-based model, which transforms different parts in the joint structure into springs with specific stiffness, has been widely used in the literature due to its simplicity to implement\cite{mccarthy2011analytical}. In the meantime, a new requirement for the spring-based model has been put forward with regard to the importance of considering the effect of friction, the relevant parameters of which include bolt-hole clearance and tightening torques. Instead of using a constant spring stiffness, the load-displacement curve shows a four-stage shape considering the transition of contact force from static frictions to sliding frictions, making a variable spring stiffness in the model\cite{qiu2019improved, mccarthy2006simple,mccarthy2005three}. While researchers have clarified influencing factors of bolt load distribution, optimization of an even-load-distribution design has seldom been focused on. 

Machine Learning, which discerns hidden patterns from complex dataset has been proven to be a very efficient tool in many research fields including solid mechanics\cite{schmidt2019recent,chen2019machine,ramprasad2017machine,zobeiry2020theory}. As a branch of artificial intelligence, on one hand, it can be incorporated as an alternative way of generating mechanical response of a given structure or compounds which used to obtain by advanced computational techniques like density functional theory, molecular dynamics, and finite element method\cite{ward2016general,huber2018machine,yang2018deep}. On the other hand, it is claimed to be a promising way to accelerate the material design process which involves an inverse problem of designing new materials or structures with desired property\cite{ma2020accelerated, liu2015predictive, chen2020generative}. Some interesting and meaningful applications includes designing hierarchical structures with high fracture toughness\cite{gu2018bioinspired}, predicting residual strength from acoustic emission data\cite{ramasamy2014prediction}, solving inverse structure-process problem in addictive manufacturing\cite{tran2020active}.

In this work, we aim to provide an even-load-distribution design process for multi-bolt CFRP composite joint using a machine learning based framework. Based on the spring model, we put forward a simple equivalent circuit model in Matlab/Simulink that can easily produce dataset for the training of machine learning model. The bolt-hole clearance and tightening torques are chosen as design inputs due to their significant influence on load distribution as well as can be easily manipulated in practice. A symbol of load distribution unevenness is defined as output to represent the level of differences in bolt loads before failure. The artificial neural network is trained as a response predictor and then used to generate a database for all the possible inputs in our design space. Finally, the optimal parameters provided by this general machine learning-based framework is validated for its reducing the load distribution unevenness in a three-bolt joint.

\section{Problem description}
In this section, the load distribution problem of a typical single-lap multi-bolted joint is introduced. The corresponding circuit model using Matlab/Simulink is established in which bolt-hole clearance and frictions caused by the tightening torque can be taken into consideration. After a brief introduction of the circuit model, the effectiveness and accuracy of the model are validated through a comparison with FEM.

\subsection{Circuit model for predicting bolt-load distribution}

\begin{figure}[ht]
	\centering
	\includegraphics[width=15cm]{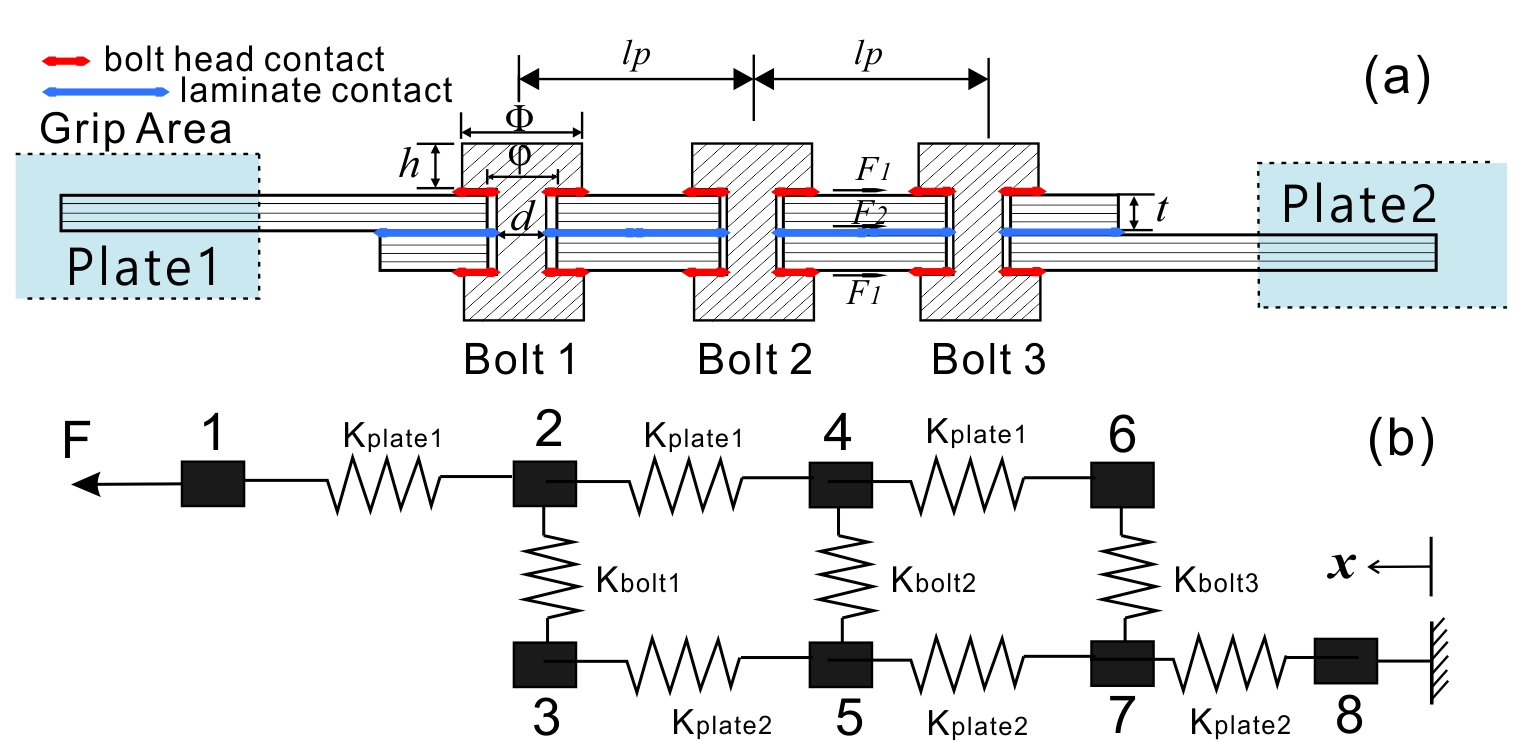}
	\caption{(a) A typical structure of three-bolt, single-lap joint (b) Spring-mass model of multi-bolt joint}
	\label{F1}
\end{figure}

For the load-sharing analysis of multi-bolted joint structure as shown in Fig.\ref{F1}, the spring-mass model is the most commonly used method which holds the advantage of low computational cost and high efficiency against other methods like FEM, analytical method, and experimental measurement. With the increasing number of bolts, obtaining bolt load from the spring-mass model becomes complicated especially when various bolt-hole clearances and tightening torques are considered, which may involve non-linear stages. To solve this problem, in our previous work\cite{qiu2019improved}, a novel circuit model is established for determining the bolt-load distribution and joint stiffness.  Then, in the circuit model, bolt load distribution is reflected by the flow of current. Here, a brief introduction of this circuit model is given.

The spring-based method, which is a stiffness method essentially, separating the global deformation under external load into several extensions of different parts. These parts are assumed to have a simple load-deformation relationship which can be represented as springs in the model. The stiffness of each spring is the key parameter, especially for the $K_{bolt}$ shown in Fig.\ref{F1} denoting the load-deformation relationship in the bolt area.

$K_{plate}$ in the spring-based model, which is the tensile stiffness of the jointed plate, can be easily calculated as,
\begin{equation}
K_{plate} = \frac{E_{px}A_{p}}{l_{p}}
\label{E1}
\end{equation}
where $E_{px}$ is the laminate modulus along the applied load direction. $A_{p}$ stands for cross-sectional area of the plate while $l_{p}$ is the bolt pitch.

It is revealed that with the introduction of bolt-hole clearance, a four-stage curve of load-displacement behavior occurs for the single-lap joint. In the beginning, when the applied load is too small to invoke any sliding, the external load is counteracted by the static frictions from the upper and lower contact surfaces on plates(which is highlighted by red and blue color in Fig.\ref{F1} ). In the second stage, as the applied load increases, the contact forces would exceed the maximum static friction, causing the slides at the contact area. This sliding is proven to take place first at the contact surfaces between two jointed plates since that ratio of $F_{1}$ to $F_{2}$ is greater than 1 shown as,
\begin{equation}
R=\frac{F_{2}}{F_{1}}=(\frac{64t^{2}}{3d^{2}(1+\mu_{b})}+2.4)\cdot \frac{A_{0}G_{p}}{A_{b}G_{b}}+1
\label{E2}
\end{equation}
Here $d$,  $\mu_{b}$, $A_{b}$ and $G_{b}$ are the diameter, Poisson's ratio, minimum cross-sectional area and shear modulus of bolt respectively. Whilst, $A_{0}$ is the contact area between bolt head and plate.  $G_{p}$ is the shear modulus of laminate.

When the increasing external load reaches the summation of maximum static frictions on both upper and lower contact surfaces,  a global sliding happens to lead to a plateau phase is shown in the load-displacement curve. Once the bolt rod contacts with the edge of bolt hole on laminates, the load-displacement curve goes into the fourth stage in which the joint stiffness keeps a constant value until final failure.

According to the above analysis, joint stiffness would vary in each of the four stages as a result of the changes in friction condition, that is,

(1) Phase 1 ($F_{2}<f_{static}$)
\begin{equation}
\frac{1}{K_{bolt}^{(1)}}=(\frac{12t}{5A_{b}G_{b}}+\frac{8t^{3}}{3E_{b}I_{b}}+\frac{(3-R)t}{4A_{0}G_{p}})\cdot\frac{1}{1+R}+\frac{(4l_{p}-3\Phi)t^{2}}{32E_{px}I_{p}}
\label{E3}
\end{equation}

(2) Phase 2 ($F_{1}<f_{static}=F_{2}$)
\begin{equation}
\frac{1}{K_{bolt}^{(2)}}=(\frac{12t}{5A_{b}G_{b}}+\frac{8t^{3}}{3E_{b}I_{b}}+\frac{3t}{4A_{0}G_{p}})+\frac{(4l_{p}-3\Phi)t^{2}}{32E_{px}I_{p}}
\label{E4}
\end{equation}

(3) Phase 3 ($F_{1}=f_{static}=F_{2}$)
\begin{equation}
\frac{1}{K_{bolt}^{(3)}}=\infty
\label{E5}
\end{equation}

(4) Phase 4 ($displacement>clearance$)
\begin{equation}
\frac{1}{K_{bolt}^{(4)}}=\frac{4t}{3G_{p}A_{p}}+[\frac{4}{tE_{b}}+\frac{2}{t(\sqrt{E_{px}E_{py}})}](1+3\beta)
\label{E6}
\end{equation}
In these equations, $t$, $\Phi$ are geometric parameters can be found in Fig.\ref{F1}. $E_{b}I_{b}$ is the bending stiffness of bolt rod. While bending moment in single-lap joints can be reacted by both laminates and bolts, the term $\beta$ here represents the fraction of the bending moment which is reacted by the non-uniform contact stresses in laminates. Here $\beta=0.15$ for a typical protruded-bolt joint. Detailed explanation for all the above equations can be found in Ref\cite{qiu2019improved}.

\begin{figure}[ht]
	\centering
	\includegraphics[width=14cm]{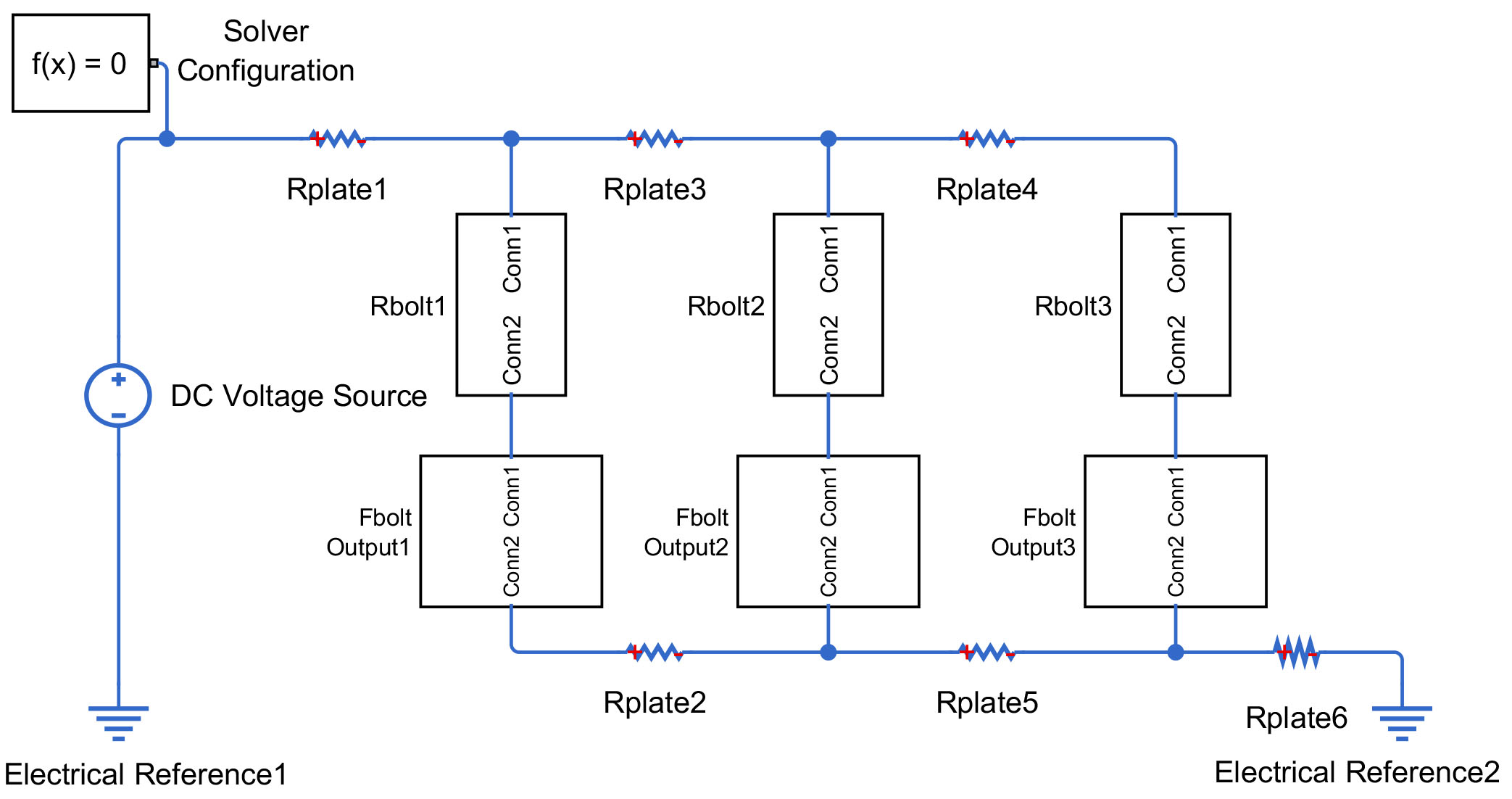}
	\caption{The circuit model for bolt load sharing analysis of a single-lap, three-bolt joint}
	\label{F2}
\end{figure}

After figuring out the corresponding stiffness of each spring, the next step is to acquire bolt load, global deformation, and global joint stiffness from spring-based model. In order to solve this problem, a novel circuit model is proposed using Matlab/Simulink. A basic model for the three-bolt joint is shown in Fig.\ref{F2}. In this circuit model, the applied voltage equals to the displacement at the loading end. Then the resistance value of each resistor is set to be reciprocal of the corresponding spring stiffness expressed in Eq.\ref{E1}-\ref{E6}, making the flow of current equals to bolt load distribution. The Matlab/simulink provides a simple way to simulate and run the whole process. Then, results can be outputted to Matlab workspace for the following post-processing and visualization work. 

\begin{figure}[ht]
	\centering
	\includegraphics[width=14cm]{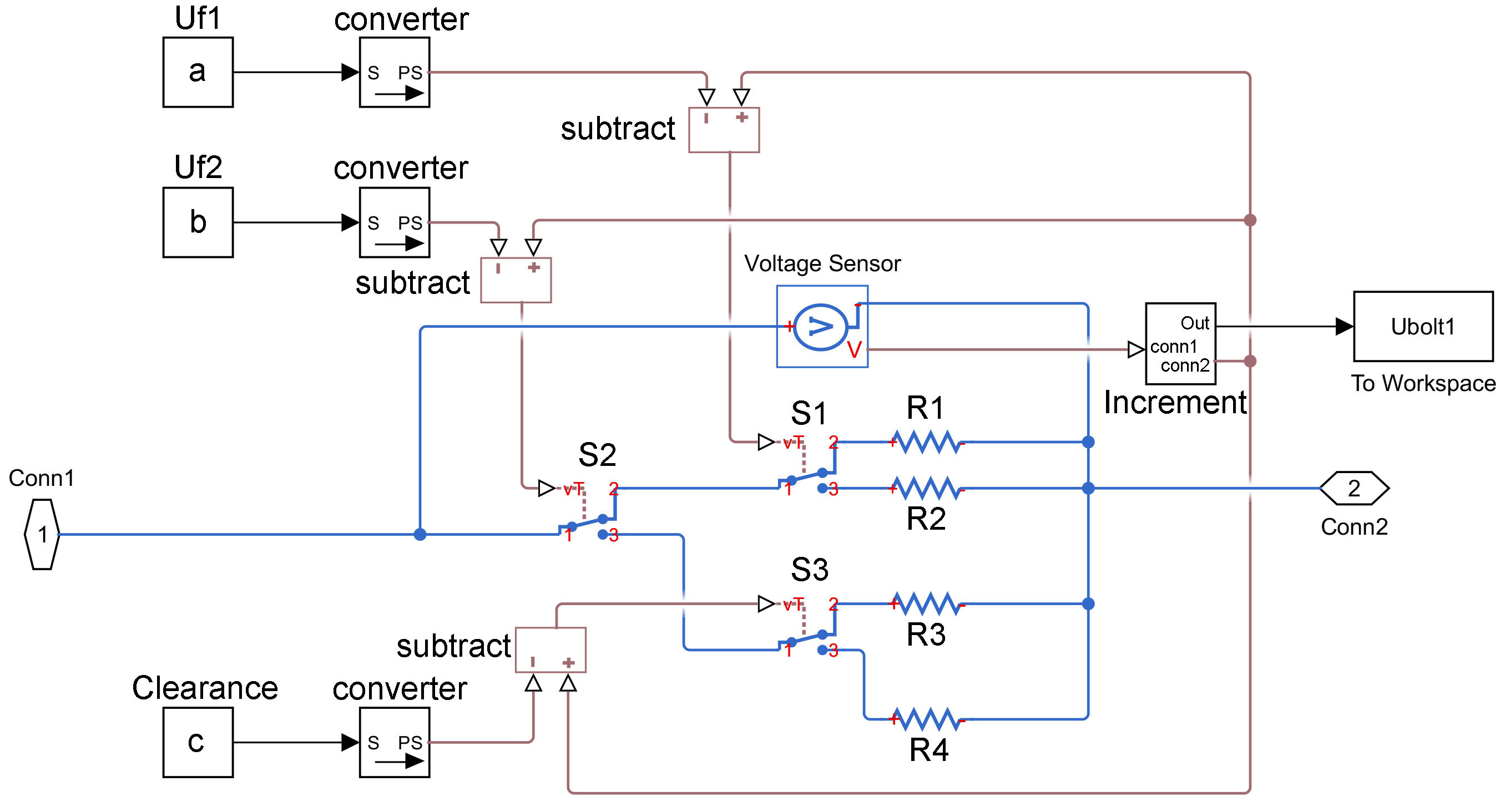}
	\caption{Module for bolt stiffness considering the effect of bolt-hole clearances and tightening torques}
	\label{F3}
\end{figure}

Special attention should be paid to the ``$R_{bolt}$" module. The established module for bolt stiffness is shown in Fig.\ref{F3}. A total of three conditional switches are incorporated to evoke the three knee points in the four-stage load-displacement curve. While the transition conditions are presented in the form of force for the first two knee points in Eq.\ref{E3} and Eq.\ref{E4}, the form of displacement which corresponds to the voltage in circuit model is adopted in this module. In practice, after measuring the voltage value using a sensor block, its value is extracted and compared to ``a", ``b" and ``c" to control the switches. ``a", ``b" and ``c" denoting for the three displacements of transition conditions are defined and imported into the circuit model using a Matlab script. Finally, the bolt load is measured by a current sensor block and outputted to Matlab workspace by the output modules shown in Fig.\ref{F2}.

The effect of bolt-hole clearances and tightening torques can be reflected by the values of ``a", ``b" and ``c". Among others, the value of ``c" here accounts for bolt-hole clearance. According to the Coulomb's friction law, the maximum static friction is proportional to tightening force which relates to tightening torques and geometries of the bolt. Thus, the displacement-control knee points can be calculated as,
\begin{equation}
a=\frac{f_{static}(1/R+1)}{K_{bolt}^{(1)}}=\frac{vT}{kd}\cdot \frac{(1/R+1)}{K_{bolt}^{(1)}}
\label{E7}
\end{equation}
\begin{equation}
b=a+\frac{f_{static}(1-1/R)}{K_{bolt}^{(2)}}=a+\frac{vT}{kd}\cdot \frac{(1-1/R)}{K_{bolt}^{(2)}}
\label{E8}
\end{equation}

Here, $v$ is the friction coefficient between contact surfaces. $T$ is the applied tightening torque and $k$ is a constant value accounting the relation between bolt torque and tightening force. $k=0.2$ and $v=0.3$ is adopted in this paper according to Ref\cite{mccarthy2005experiences}.

\subsection{Validation of circuit model}

A three-bolted joint with its geometries shown in Tab.\ref{T1} is studied in this paper. The composite layup is $[+45/0/-45/90]_{3S}$ for both laminates, whose nominal thickness is 3 mm. The lamina properties are $E_{1}=195.00$ GPa, $E_{2}=8.58$ GPa, $G_{12}=4.57$ GPa. The homogeneous mechanical properties of laminate are computed using the Classical Laminate Theory and listed in Tab.\ref{T1} as well as the mechanical parameters of the bolt. 

In the circuit model, the resistors marked as ``$R_{plate}$" are assigned with a value reciprocal to Eq.\ref{E1} while the resistors in the ``$R_{bolt}$'' module have a value that corresponds to Eq.\ref{E3}-Eq.\ref{E6} respectively. It has to be noted that a fixed-step solver option is adopted in Matlab/Simulink. Key parameters in solver control include value of the provided Voltage Source $V_{s}$, single step time $t_{0}$ and total simulation time $t_{s}$. With these parameters, the total simulated displacement can be expressed as,
\begin{equation}
V_{total}=\frac{t_{s}}{t_{0}}\cdot V_{s}
\label{E9}
\end{equation}

In practice, for each step, the Voltage Source provides a  0.005V voltage increment. With the combination of solver settings for analysis time  60s and fixed-step time 0.1s, the total applied voltage is 3V, that is 3mm displacement applied at the loading end.

\begin{figure}[ht]
	\centering
	\includegraphics[width=14cm]{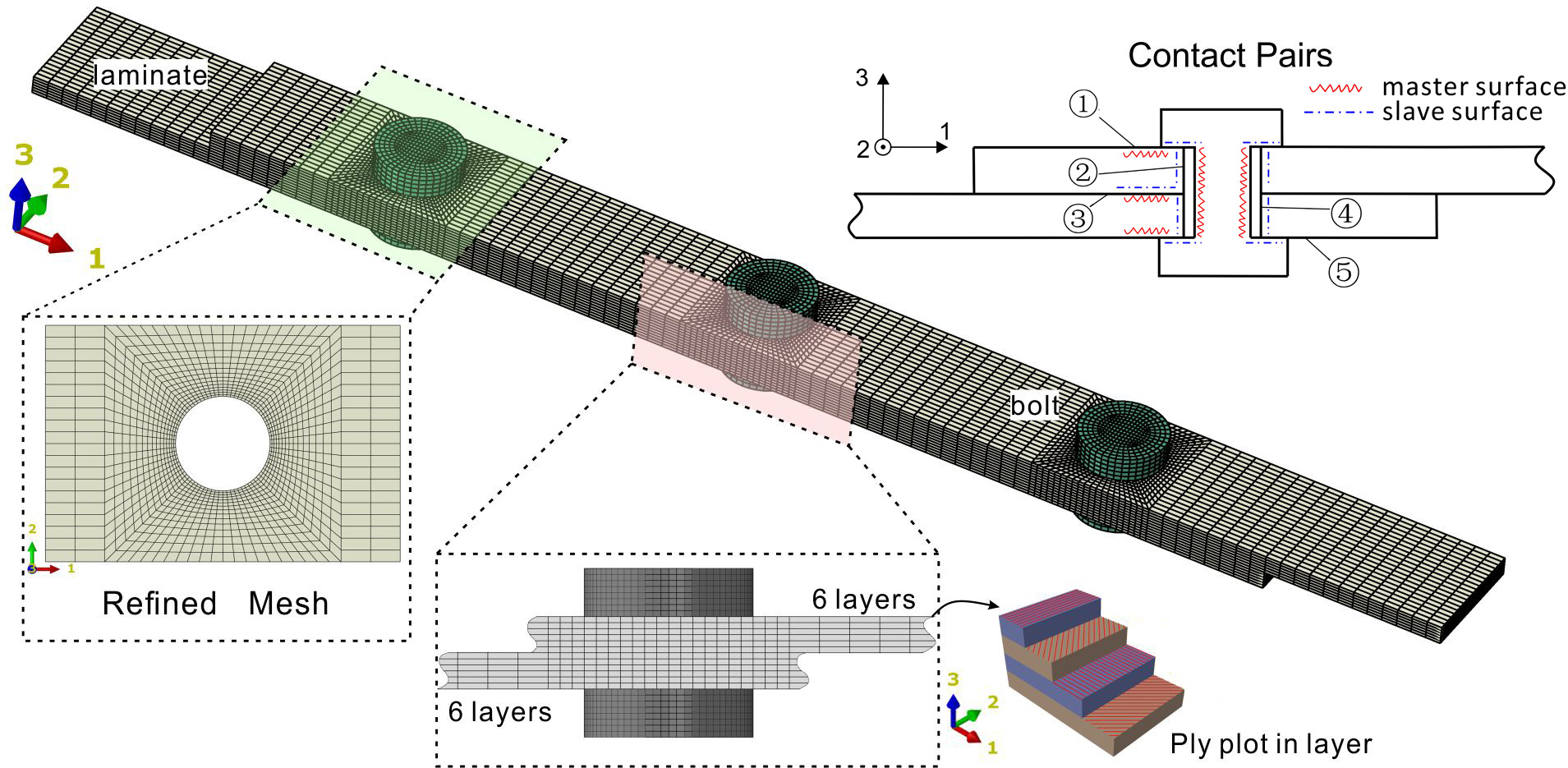}
	\caption{3D finite element model of a three-bolt joint containing 5 contact pairs}
	\label{F4}
\end{figure}

A three-dimensional numerical model is established in Abaqus software to provide FEM results as a reference to our circuit model. All of the five parts contained in the numerical model are 3D solid with reduced-integration, 8-node linear brick element C3D8R, as illustrated in Fig.\ref{F4}. A radiation-shape mesh with a bias ratio of 5 is adopted to refine mesh size near each bolt hole.

While the three bolts are assigned with a homogeneous isotropic material property, each of the laminate is partitioned into 6 layers along its thickness in which each layer represents $[+45/0/-45/90]$ ply in laminates. Then, the lamina properties are assigned to each ply along with a local material orientation according to its ply angle.

A total of 5 contact pairs are included in the model using a surface-to-surface contact method as shown in the figure. According to Abaqus user's manual\cite{abaqus2007abaqus}, surfaces with coarse mesh size and greater material stiffness are more suitable to set as the master surfaces to prevent penetration between adjacent surfaces. While the default ``Hard'' contact is used for the normal contact property, a penalty method with friction coefficient 0.3 is defined to simulate tangential contact behavior. All the degrees of freedom of both ends of the assembled structure are coupled to a controlling point to apply displacement and constraints.

The pre-tightening force is introduced by using an expansion method. As listed in Tab.\ref{T1}, the bolts are assigned with three thermodynamic parameters $\alpha_{1}, \alpha_{2}, \alpha_{3}$ which represent the coefficient of thermal expansion along local 1, 2, and 3 direction respectively. The $\alpha_{1}$ and  $\alpha_{2}$ are set to be zero to make sure that there is no other contact force except for the pre-tightening force throughout this temperature drop process. It is noted that the value of $\alpha_{3}$ and also the temperature change are chosen and adjusted to make the normal contact force between laminates and bolt head to be the desired pre-tightening force. A two-step calculation is conducted in Abaqus, in which the pre-tightening force is introduced by a temperature drop in Step 1 while the tensile loading is applied in Step 2. Here in this model, a temperature drop of $60^{o}$ is defined in this step to generate a normal contact force of  4.3 kN, which corresponds to a tightening torque of 7 Nm.

\begin{table}[!h]
	\centering
	\renewcommand*{\arraystretch}{1.2}
	\caption{Geometries and mechanical properties of a three-bolt  composite joint}
	\begin{tabular}{m{2.5cm}<{\centering} m{2cm}<{\centering} m{2cm}<{\centering} m{2cm}<{\centering} m{1.5cm}<{\centering} m{2.5cm}<{\centering} m{1.5cm}<{\centering}}\toprule
		\label{T1}
		Geometries & $l_{p} $(mm) & $d $ (mm)& $\Phi $(mm) & h(mm) & width (mm) &\\
		Value & 60 & 8 & 14 & 8 &20&\\
		\hline
		Laminate &$E_{px} $(GPa)&$E_{py} $(GPa)&$G_{p}$(GPa)&    & &\\
		Value & 71.60 & 71.60 &  4.57 & & & \\
		\hline
		Bolt  &$E_{b}$ (GPa)&$G_{b} $ (GPa)&$\mu_{b}$& $\alpha_{1} (/^{o}C)$& $\alpha_{2} (/^{o}C)$&$\alpha_{3} (/^{o}C)$\\
		Value &109.78&41.27&0.33&0&0&0.0001\\
		\bottomrule
	\end{tabular}
\end{table}

First, a three-bolt joint with uniform bolt-hole clearances and tightening torques is studied by both FEM and our circuit model. The bolt-hole clearance is 0.1 mm which is acquired by slightly reducing the radius of the bolt. And the tightening torques are equally to be 7 Nm, acquired by adjusting the expansion parameters of bolt according to the above analysis. 

\begin{figure}[ht]
	\centering
	\includegraphics[width=16cm]{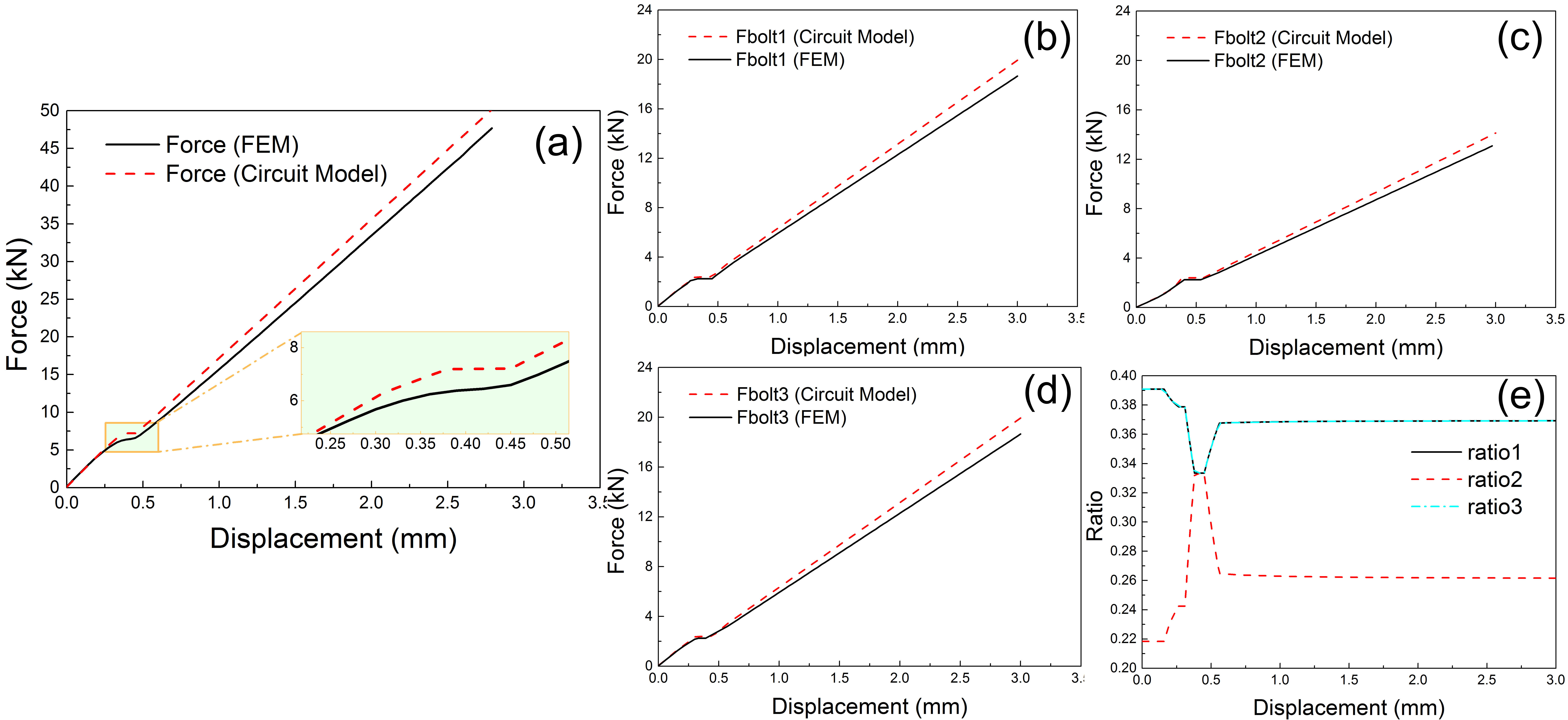}
	\caption{Comparison between results by FEM and circuit model (a)Load-displacement curve of the whole structure  (b)(c)(d)Bolt load shared by each bolt vs. global displacement (e)Load distribution ratio of three bolts vs. global displacement}
	\label{F5}
\end{figure}

Comparison between results by FEM and circuit model is shown in Fig.\ref{F5}. A similar trend is shown by the load-displacement curves of both models illustrated in Fig.\ref{F5}(a). In the beginning, when the applied load is quite small, the overall stiffness is determined by the deformation of bolt and laminates under friction. A good agreement is found between circuit model and FEM which represents the accuracy of Eq.\ref{E3} in dealing with structural deformation under friction. As the load increases, sliding firstly comes out in bolt 1 and bolt 3, due to the fact that bolt load shared by bolt 1 and bolt 3 is greater than the central bolt from the beginning. Then, this changed stiffness of bolt 1 and bolt 3 would affect the overall bolt load distribution, causing more time for bolt 2 to enter into the sliding phase. While the transition on the load-displacement curve caused by the stiffness change is a piecewise form for the results of circuit model, a rather smooth curve in shown by the curve of FEM. One of the reasons may be the more delicate consideration of contact by FEM such as the change of contact area, the non-uniform contact force, and the slight rotation of contact direction. Another reason can be the fewer data points for fitting the curve by FEM. As the load continues to increase, full contact between bolt rod and hole is achieved. A good agreement can also be found in the curves of three bolt loads, as shown in Fig.\ref{F5}(b)-(d).

Bolt load distribution ratio predicted by the circuit model is shown in Fig.\ref{F5}(e). As a result of the symmetry in three-bolt joint, bolt loads shared by bolt 1 and bolt 3 are exactly the same, which are greater than load shared by bolt 2. Several platform stages are undertaken in the beginning due to the transition of the contact status at the bolt area. After that, it is shown that in this condition of uniform bolt-hole clearance and tightening torque, the maximum ratio is 37.2\% while the minimum ratio is 25.6\%. In this case, when the total appiled load is 50 kN, a difference of 5.8 kN can take place between bolts.

\subsection{Effect of the randomness of clearances and tightening torques}

In the last section, the effectiveness of our circuit model for the load distribution calculation is validated by FEM. Here, the effect of the randomness of clearances and tightening torques is investigated. Since that the circuit model is built in Matlab/Simulink, random parameters are produced by a Matlab script and exported into the circuit model. For the clearances, a random number ranges from 0 to 2 mm is chosen for each bolt, while the random tightening torques range from a hand-tight 0.5 Nm to a highly-tighten torque 15 Nm. 

\begin{figure}[ht]
	\centering
	\includegraphics[width=16cm]{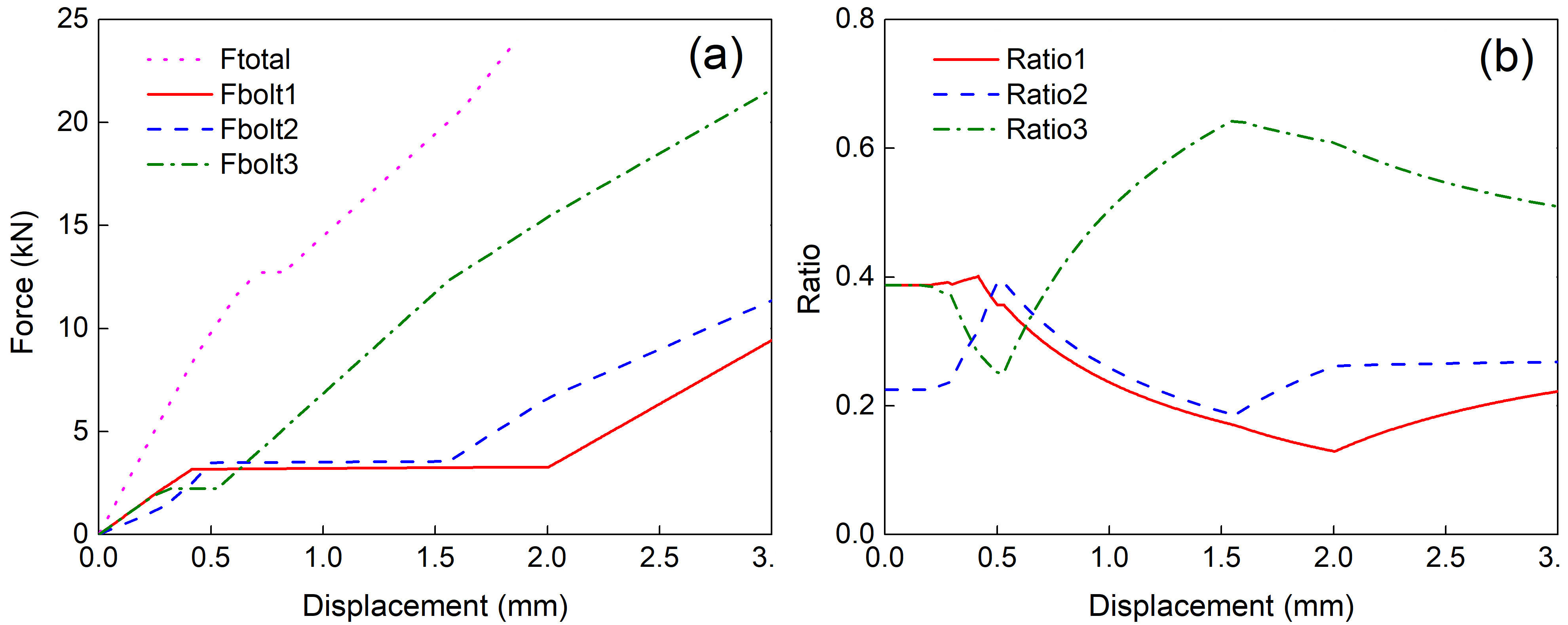}
	\caption{Results of the three-bolt joint with random bolt parameters of clearances (1.54mm, 0.63mm, 0.12mm) and tightening torques (8.41Nm, 9.25Nm, 14.90Nm) (a)Bolt load-displacement curve (b)Bolt load ratio}
	\label{F6}
\end{figure}

In the case shown in Fig.\ref{F6}, the parameters for bolt-hole clearances are (bhc1=1.04mm, bhc2=0.63mm, bhc3=0.12mm) and the torques are (T1=8.41Nm, T2=9.25Nm, T3=5.90Nm). Compared to Fig.\ref{F5} where the parameters are identical, the bolt load-displacement during the loading process has significantly been affected. In Fig.\ref{F5}, the load shared by bolt 1 and bolt 3 is equal. With the randomness of the clearances and torques, this equality is changed. In the case of Fig.\ref{F6}, bolt-hole clearance bolt 3 is smaller compared to other two bolts, resulting in a very quick jump into the full contact phase. As stated before, the changing stiffness of bolt 3 affects the overall load distribution of joint, causing a different transition point and duration of the plateau on load-displacement curves of the other two bolts. Moreover, it can be seen that the total applied force illustrated in Fig.\ref{F6}(a)  does not show an obvious piece-wise shape caused by the friction effect, while in fact a highly uneven load distribution is attained within joint. Load distribution ratios presented in Fig.\ref{F6} show a complicated pattern where the maximum ratio can be up to 64.1\%. However, since a constant stiffness is achieved if all three bolts turn into full contact, a converged value of load ratio could be obtained with increasing external load. Results suggest that at the displacement of 3mm, the maximum ratio changes from the beginning value 64.1\% to a value around 50\%, while the minimum ratio changes from the beginning value 13\% to a final value around 22\%.

Comparison of the bolt load distribution ratio and bolt load (at total load =30 kN) in the three-bolt joint with identical and random bolt parameters is shown in Tab.\ref{T2}. It can be seen that, with the varied clearances and torques, load distribution among these three bolts can be affected. Therefore, this effect of various parameters inspires us to find a set of parameters that can help to narrow down the differences in the load distribution.

\begin{table}[!h]
	\centering
	\renewcommand*{\arraystretch}{1.2}
	\caption{Comparison of bolt load distribution ratio between identical and random bolt parameters under a total load of 30 kN}
	\begin{tabular}{m{1.5cm}<{\centering} m{6.5cm}<{\centering}m{2cm}<{\centering} m{2cm}<{\centering} m{2cm}<{\centering} }\toprule
		\label{T2}
		Cases & Parameters (bhc1, bhc2, bhc3, T1, T2, T3) &Bolt 1&  Bolt 2& Bolt 3 \\
		\hline
		Identical & (0.1, 0.1, 0.1, 7.0, 7.0, 7.0) & 11.2KN (37.2\%)&  7.7 kN (25.6\%) &  11.2 kN (37.2\%) \\
		Random & (1.04, 0.63, 0.12, 8.41, 9.25, 5.90) & 4.98 kN (16.6\%)&  7.94 kN (26.5\%) &  17.08 kN (56.9\%) \\
		\bottomrule
	\end{tabular}
\end{table}

\section{Neural network for even-load-distribution design}

In this section, by adopting the back-propagation neural network, we are trying to form a tool for designing a multi-bolt joint having an even-load-distribution under the common tensile load.

\begin{figure}[ht]
	\centering
	\includegraphics[width=17cm]{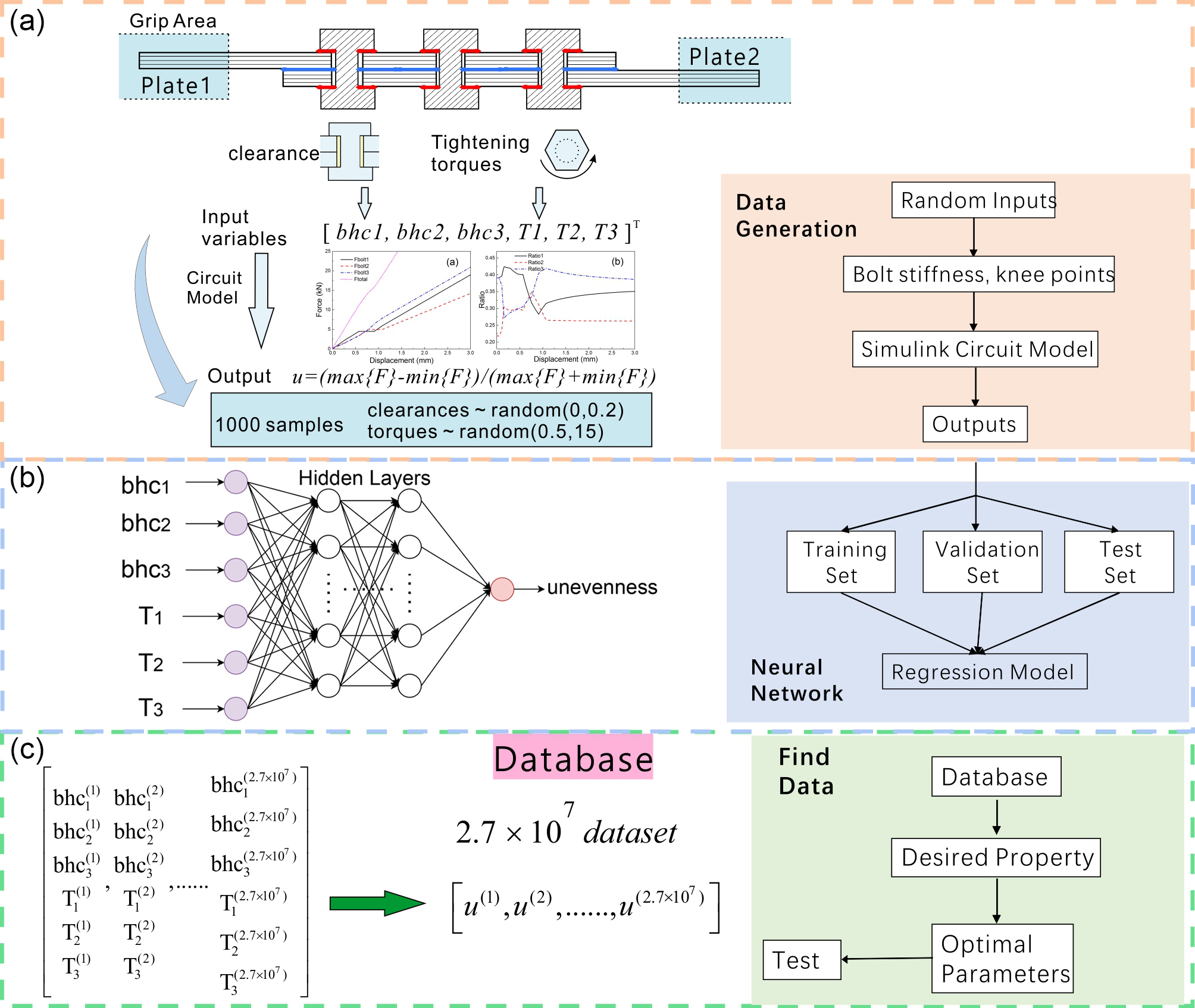}
	\caption{Machine-learning based framework for a even-load-distribution design of three-bolt joint (a)Data Generation: using circuit model in Matlab to generate 1000 samples (b)Neural Network: training and validation of the established neural network (c)Find Data: using neural network to form a database and find desired data of minimum symbol of uneven load distribution}
	\label{F7}
\end{figure}

The general framework of this machine learning-based optimization work can be summarized by Fig.\ref{F7}. It can be seen that our work can be divided into three parts, that is ``Data Generation", ``Neural Network", and ``Find Data". First, the input variables include three bolt-hole clearances and three tightening torques. As explained before, the design space for bolt-hole clearance is between 0 to 2 mm and for tightening torque is between 0.5 to 15 Nm. Here, an increment size of 0.2 mm is used for clearances which makes 1000 possible combinations for three-bolt clearances. In the meanwhile, an increment size of 0.5 Nm is adopted for torques which makes 27000 possible combinations for three-bolt torques. Thus, a total of $2.7\times 10^{7}$ possible test patterns are assumed. The following symbol for the evaluation of unevenness is adopted as outputs:
\begin{equation}
u=\frac{max\{F_{i}\}-min\{F_{i}\}}{max\{F_{i}\}+min\{F_{i}\}}\quad (i\ for\ total\ load = 30 kN)
\label{E10}
\end{equation}

Here, $max\{F_{i}\}$ indicates the maximum bolt load when to total applied load is 30kN while $min\{F_{i}\}$ indicates the minimum bolt load. If the unevenness of load distribution is low, this $u$ value should be small.

In Data Generation, 1000 random inputs and corresponding outputs are generated by Matlab script using our circuit model. In practice, this step takes about 20 minutes mainly due to the call and calculation of the circuit model in Matlab/Simulink. Then, these 1000 samples are used for the training process of our back-propagation neural network. It is noted that the 1000 samples are divided into training set, validation set, and test set using a ratio of 70\%/10\%/20\%. The architecture of the established neural network is shown in Fig.\ref{F7}. Four hidden layers are included and the number of neurons in each layer is 30/40/40/30 respectively. The stochastic gradient descent back-propagation optimizer ``Adam" is used as the learning algorithm while the rectified linear function (ReLU) is used as the activate function for calculation between layers. After the training and evaluation of neural network, it is used as a response predictor to obtain a symbol of the unevenness of the test pattern.  A database is thus formed by these test patterns and their predictions. With this database, a script of pattern generator is written to get the optimal parameters of bolt-hole clearances and tightening torques which make the smallest symbol of distribution unevenness. Finally, the optimal results are tested and evaluated by comparing with FEM and results by the circuit model.

\begin{figure}[ht]
	\centering
	\includegraphics[width=16cm]{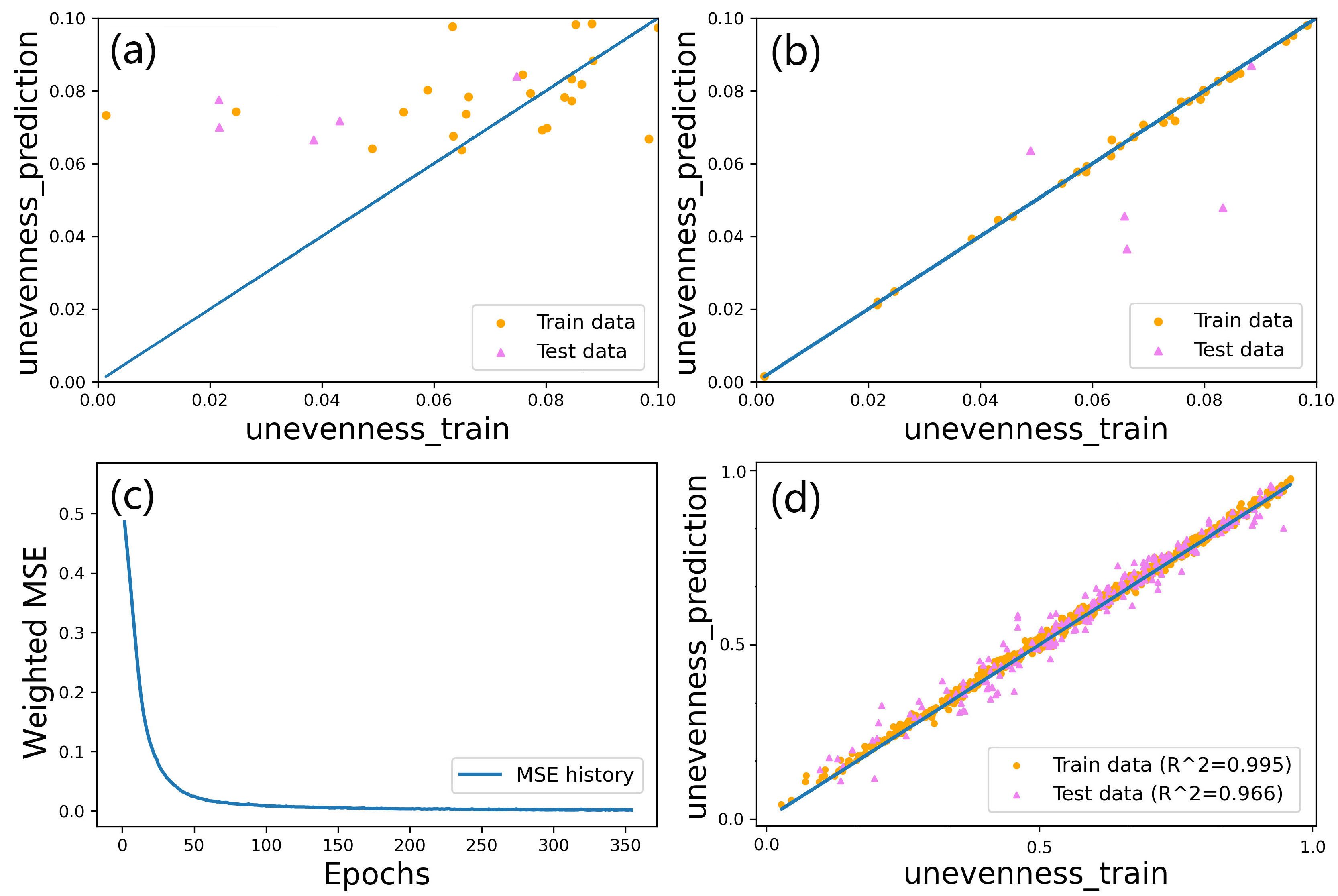}
	\caption{Predicting accuracy with target values in the range (0, 0.1) (a)normal MSE (b)Weighted MSE; Performance of the neural network model (c)History of MSE with increasing epoch number (d)Regression coefficient for training and test dataset}
	\label{F8}
\end{figure}

The ultimate goal of a neural network is to find a group of weights and bias for every neurons to make a minimum loss function, which indicates the difference between predictions and target values. Here, a weighted mean square error (MSE) is defined as loss function to increase the accuracy of predicting a relatively low value of unevenness $u$, which is important for our optimization problem. The adopted loss function is shown in Eq.\ref{E-loss}. Here, $ y_{i}$ is the predicted value and $\hat{y_{i}}$ is the target value. Compared with the commonly used MSE, a weighting factor $1/(\hat{y_{i}}+0.001)$ is added, which will make a larger value of loss function when the target value $\hat{y_{i}}$ is smaller. With the incorporation of this customized loss function, the accuracy of our neural network is increased with $\hat{y_{i}}$ in the range (0, 0.1) as presented in Fig.\ref{F8}(a) and (b).

\begin{equation}
\label{E-loss}
\renewcommand*{\arraystretch}{1.5}
\text{Weighted MSE} =\frac{1}{n}\sum_{i=1}^{n}\frac{(y_{i}-\hat{y_{i}})^{2}}{\hat{y_{i}}+0.001}
\end{equation}

Having built the neural network architecture using a widely used Python machine-learning package ``TensorFlow", we then focus on the validation and accuracy of the response predictor. One common way to evaluate the performance of this regression model is through the loss function, i.e. the weighted MSE. A smaller loss function value means better performance. Fig.\ref{F8}(c) shows the history of the weighted MSE with increasing epoch number. Every time the epoch number increases, the whole dataset is used and the regression parameters within neural network are updated according to the chosen learning algorithm, thus making a smaller MSE if they are updated towards the right direction. It can be seen that with the increasing epoch number, MSE gradually decreases from 0.5 to the minimum MSE 0.0015. In order to avoid over-fitting, we set a stopping criterion that training process ends when the loss function MSE does not improve more than 1e-5 for 20 consecutive epochs. So the training process ends at epoch 355. Another common representation of the performance for this machine learning model is through a comparison between every single prediction and its reference value. This comparison is shown in Fig.\ref{F8}(d) in which x-axis is the reference value and y-axis is the predicted output. A closer position of a dot to the bisection line ($y=x$) suggests a more accurate predicted value. In the figure, this is also reflected by the regression coefficient "$R^{2}$" ranging from 0 to 1. The closer the value to 1, the better matching of the neural network. It can be seen that the value of R is very close to 1 for both training data and test data. And the band width of the dots is narrow, indicating a satisfying overall performance of the neural network.

After successfully  validated the robustness of neural network, it is used to generate all the corresponding predictions of all possible $2.7\times 10^{7}$ test patterns. It is noted that the precision of this database is controlled by the increment size of clearances and torques. Considering the actual assembly equipment and measurement difficulty, we think that the step size of 0.2 for clearances is acceptable. Since that the effect of tightening torques is directly reflected by the friction which is nearly one-fifth of the torque in our cases according to Eq.\ref{E7}, a larger step size of 0.5 for torques is acceptable. According to this database, the optimal parameters of the desired property, which is the minimum symbol of unevenness, are as follows:

\begin{itemize}
	\item Clearances: $bhc1= 0.4 mm, bhc2= 0.2 mm, bhc3= 0.4 mm$
	\item Torques: $T1= 11 Nm, T2= 10 Nm, T3= 15 Nm$
\end{itemize}

\begin{figure}[ht]
	\centering
	\includegraphics[width=16cm]{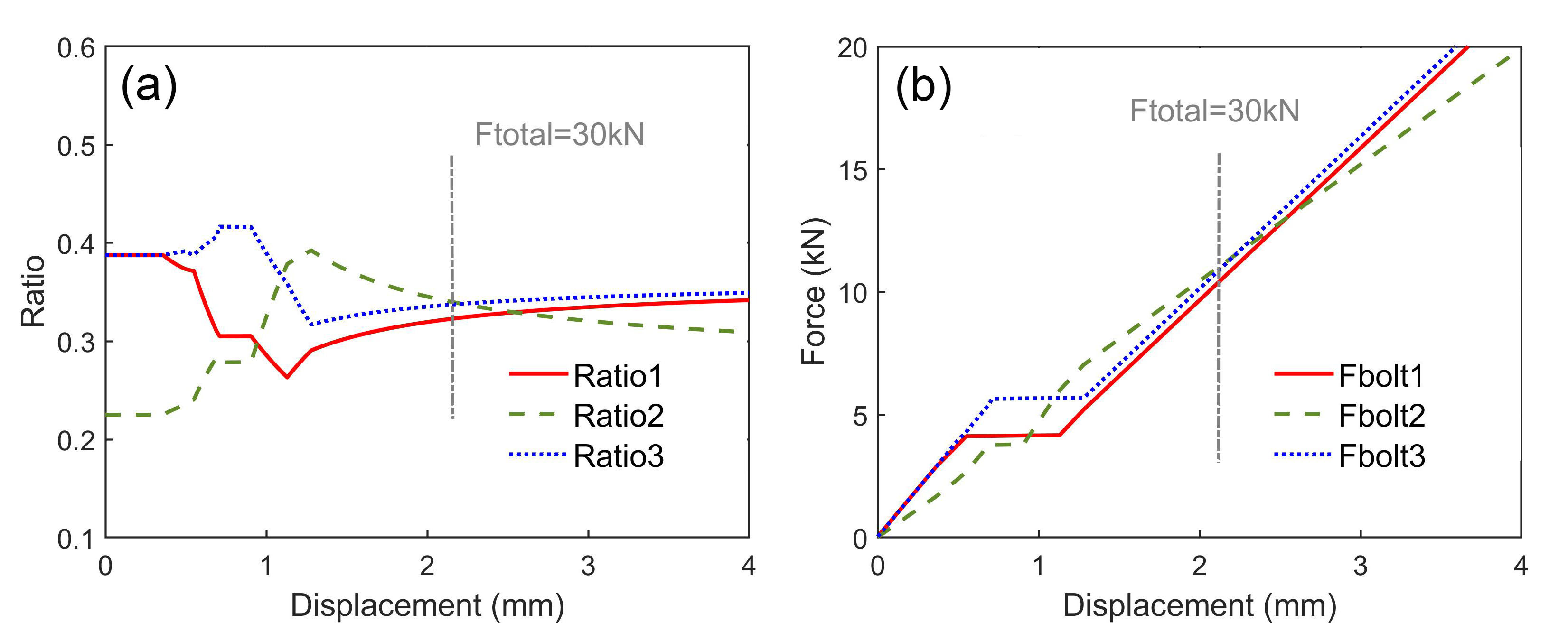}
	\caption{Loading response of three-bolt joint with optimal parameters for even-load-distribution design provided by machine learning based framework (a)Bolt load-displacement curve (b)Bolt load ratio}
	\label{F9}
\end{figure}

Since that multi-bolt joint using identical parameters illustrate a relatively low load ratio shared by bolt 2, the optimal parameters provided by neural network indicate a smaller bolt-hole clearance at bolt 2 will help to alleviate the load carried by bolt 1 and 3 while increase the level bolt 2 is occupied. The detailed reason behind this can be explained through a careful examination of the loading response of the three-bolt joint with the optimal parameters, which is shown in Fig.\ref{F9}. It can be seen that joint with these chosen parameters renders bolt 2 a high initial bolt load ratio, which is up to 41\% when the total applied load is 18.0 kN and displacement is 1.25 mm. Since that our algorithm only cares about the load distribution when the total applied load is 30kN, this high unevenness at the beginning phase is not taken into account. According to the results shown in Fig.\ref{F5}, when all three bolts enter into the phase of full-contact, the load distributed on bolt 2 is clearly less than the other two bolts.  Therefore, for results of Fig.\ref{F9}, the high initial load ratio of bolt 2 start to decrease as full contact begins. Then a low value of unevenness is successfully acquired at the total load of 30kN. The load ratio shared by each bolt is 32.1\%/34.7\%/33.2\% relatively, which shows a good even-load-distribution.

Although this machine learning-based framework can be successfully implemented to give a satisfying set of optimal parameters than can fulfill the demand of minimum unevenness, its efficiency needs to be valued. It takes almost 20 minutes for this whole design campaign of a specific multi-bolt joint. This is a huge advantages of our circuit model built in Matlab/Simulink over FEM since our FE model would take about 5 hours generating one single data. The time-consuming operation of FEM usually comes from the introduction of contact and three-dimensional elements, which may increase the difficulty of convergence. A further effort can be made by using parallel computing in Matlab, or decode the circuit model and replace it by a equal script.

\section{Discussion}
\subsection{Comparison with other optimization algorithm}
In the above work, we use this machine learning model to efficiently build a database for all the possible  combinations of design variables in the given design space, and then can directly find the best outcomes. A common way to solve the optimization problem is by using optimization algorithm like Genetic Algorithm(GA), Ant Colony Optimization(ACO), and Particle Swarm Optimization (PSO). These advanced algorithms are inspired by the nature for finding the right iteration direction given initial dataset. Here, results and computational efficiency of these algorithms, machine learning model, and also their combinations are compared.

\begin{figure}[ht]
	\centering
	\includegraphics[width=16cm]{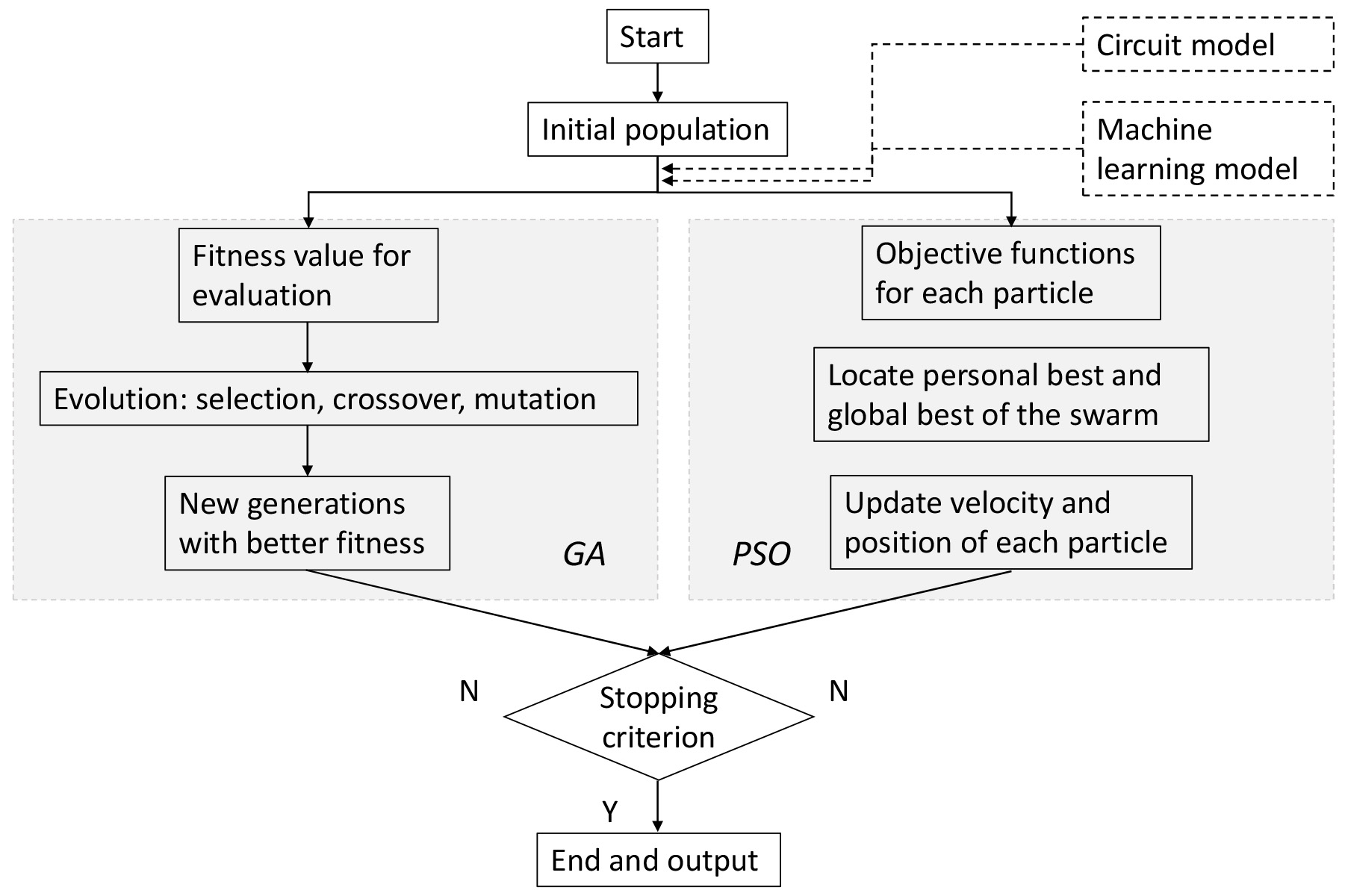}
	\caption{Flowchart of genetic algorithm (GA), particle swarm optimization algorithm (PSO), and the combination with machine learning model}
	\label{fig-GA}
\end{figure}

The work flow of the optimization process is shown in Fig.\ref{fig-GA}. Here, both Genetic Algorithm and Particle Swarm Optimization algorithm provide a way to approach minimum fitness function value at each iteration. When the stopping criterion is reached, which is typically evaluating whether the fitness function has converged, the optimization process stops and best output is reported. With the circuit model and machine learning model as the predictor of fitness value, a total of four methods are compared to our machine learning-based framework, that is, (a) GA+circuit model (b) PSO+circuit model (c) GA+machine learning model (d) PSO+machine learning model.

\begin{figure}[ht]
	\centering
	\includegraphics[width=14cm]{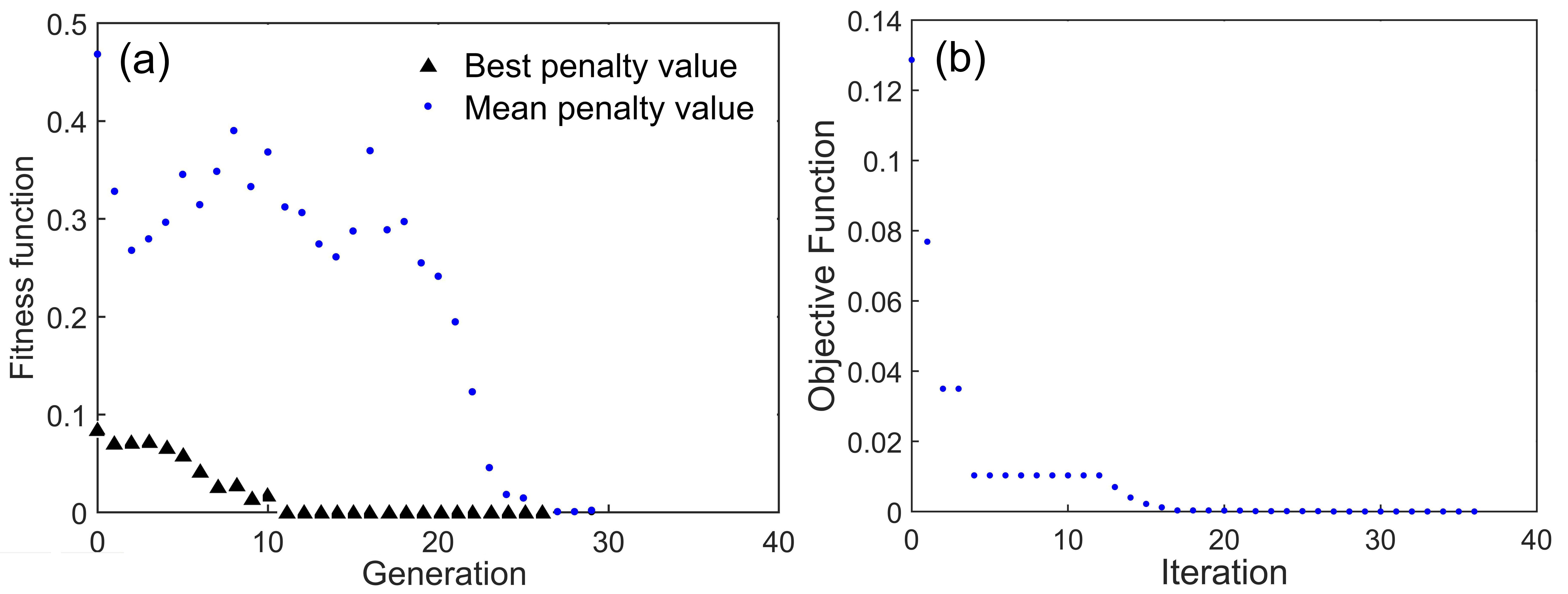}
	\caption{The optimization history curves of (a)GA and (b)PSO at each iteration}
	\label{fig-op-his}
\end{figure}

For the operation of these algorithms, stopping criterion is one of the most important factors affecting the computational efficiency. A same stopping criterion is adopted for all the optimization process, which will end the progress when the relative change in best objective function value over the last 20 iterations is less than 1e-3. The optimization history curves of GA and PSO are shown in Fig.\ref{fig-op-his}. It can be seen that both PSO and GA can quickly reach a convergence at around 13 iterations. The computational time cost listed in Tab.\ref{T3} clearly shows that, on one hand, using PSO can have slightly greater time-efficiency than GA. On other hand, it can be shown that by using machine learning model as a surrogate model for circuit model, the time cost by the optimization process can be greatly reduced. For each configuration, only 1.2 second is needed. This significant improvement is partly because of the fact that neural network algorithm is quite simple and easy to implement, and partly because the circuit model-based optimization process needs to call the Matlab/Simulink module at each iteration. It has to be noted that 20 minutes have already been spent on generating training dataset for this neural network. However, firstly the training data needed can be further reduced and secondly, the machine learning model can be expanded when designing joints with other geometric configurations.

\begin{figure}[ht]
	\centering
	\includegraphics[width=16cm]{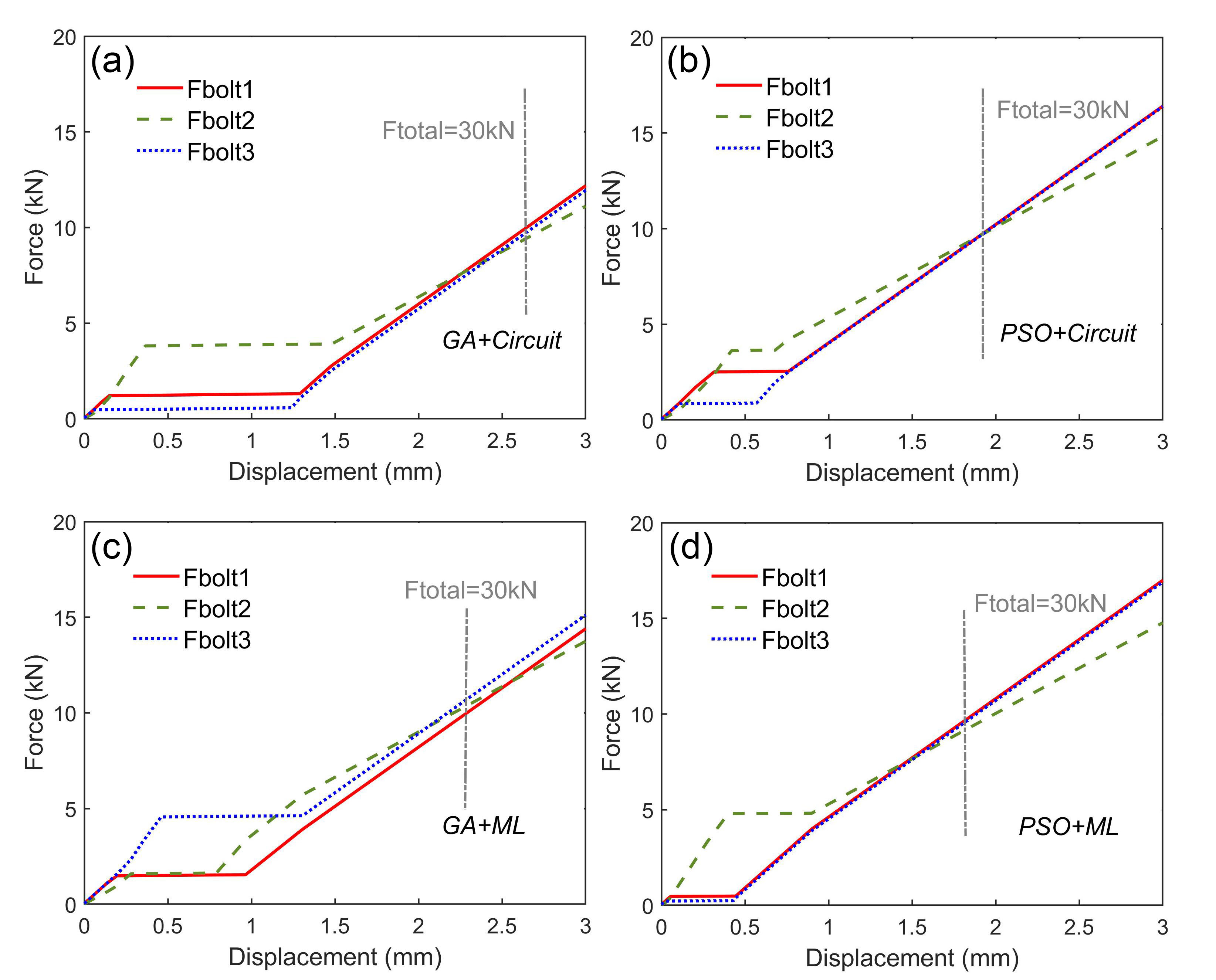}
	\caption{The bolt load -displacement curves calculated by circuit model using the optimized parameters provided by the four optimization methods: (a)GA+Circuit model, (b)PSO+Circuit model, (c)GA+Machine learning, and (d)PSO+Machine learning}
	\label{fig-result}
\end{figure}

Comparing the optimized parameters provided by these methods and also by our database using the framework shown in Fig.\ref{F7}, it can be found that the unevenness values shown in Tab.\ref{T3} are smaller than the previous results shown in Fig.\ref{F9}, indicating a better design. This is due to the prescribed precision in our database as bolt-hole clearance increases by 0.2mm while torque increases by 0.5 Nm. But using these optimization algorithms, the input precision can be set to 0.01, which greatly expand the whole design space. Therefore, better results can be found. Here, the optimized parameters listed in Tab.\ref{T3} are different by different methods. However, all of their corresponding bolt load -displacement curves  as presented in Fig.\ref{fig-result} show good evenness of bolt-load-distribution. This suggests that our optimization problem here may have local minimums, which makes many different design parameters can lead to a close value of unevenness.

\begin{table}[!h]
	\centering
	\renewcommand*{\arraystretch}{1.2}
	\caption{Comparison of results and computational efficiency of the optimization methods}
	\begin{tabular}{m{2.5cm}<{\centering} m{6cm}<{\centering} m{3cm}<{\centering} m{2.5cm}<{\centering} }\toprule
		\label{T3}
		Methods & Best input (bhc1, bhc2, bhc3, T1, T2, T3)&  Best output (unevenness \textit{u})& Time \\
		\hline
		GA+Circuit & (1.0, 0.9, 1.0, 3.2, 10, 1.2)&  0.019 &  86.7min \\
		PSO+Circuit & (0.32, 0.21, 0.28,6.6, 9.6, 2.2)&  0.007 &  67.8min \\
		GA+ML & (0.58, 0.41, 0.60, 3.9, 4.2, 12.1)&  0.031 &  9.1s \\
		PSO+ML & (0.2, 0.2, 0.2, 1.2, 12.6, 0.5)& 0.021  &  1.2s \\
		\bottomrule
	\end{tabular}
\end{table}

\subsection{Experimental study for validation}
Followed by the intuition given by the optimization results, several composite three-bolt joints were fabricated for validating the bolt load distribution. Carbon fiber/epoxy prepregs were used for fabricating, which were cured under 80 degree for 2hours. And then the cured plate was cut by a waterjet cutting machine ProtoMax with high positional accuracy of 0.05mm. For all the composite laminates, a layup of $[+45/0/-45/90]_{3S}$ is used. Their homogeneous laminate property and geometries are listed in Tab.\ref{T1}. The bolt hole diameters are different for the two groups of test specimens to generate different bolt-hole clearances when assembled with same steel bolts.

\begin{figure}[ht]
	\centering
	\includegraphics[width=12cm]{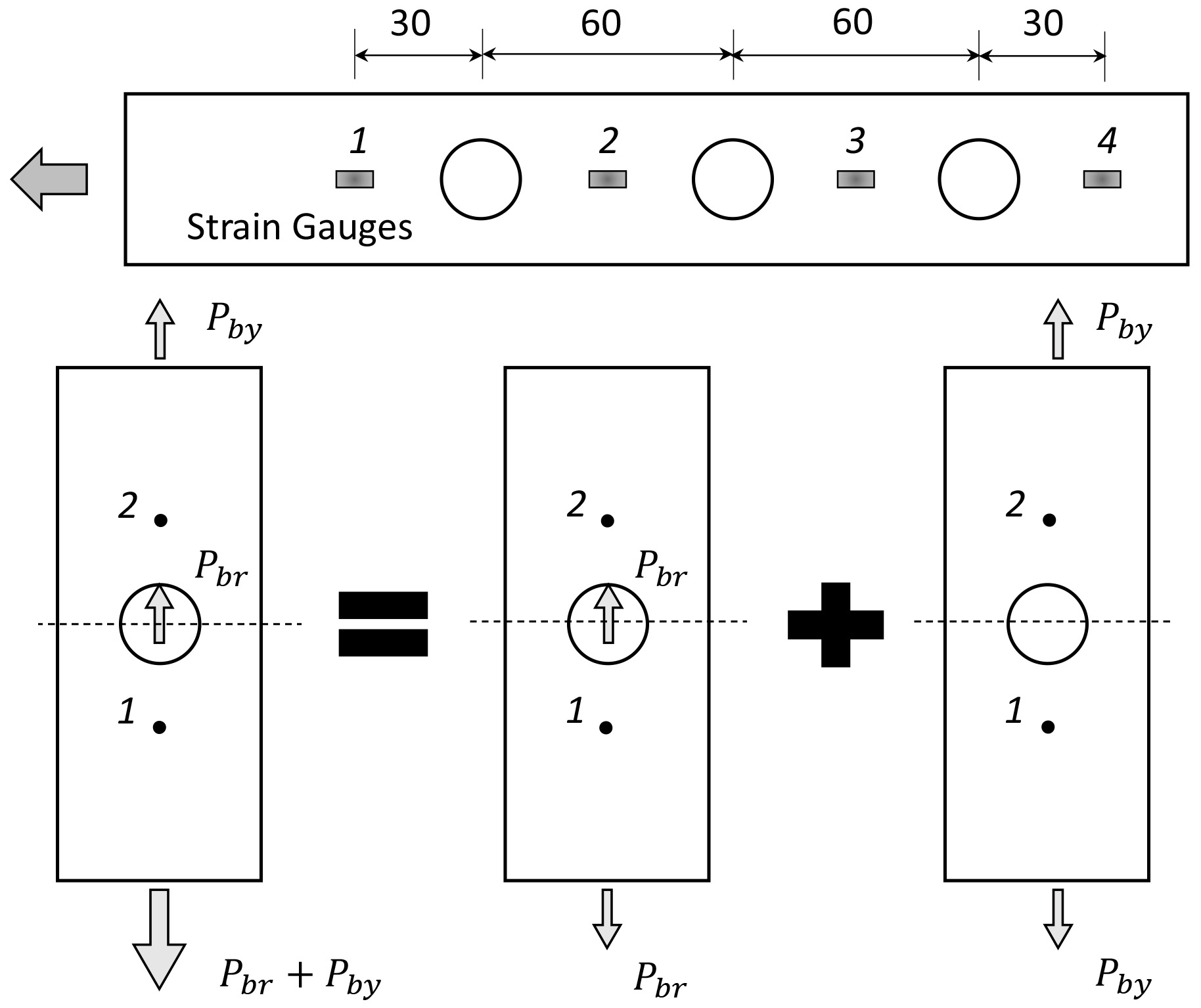}
	\caption{Strain gauges placed on the multi-bolt joint structures and the stress analysis}
	\label{fig-stress}
\end{figure}

The key problems for this experimental validation is how to evaluate the bolt load distribution in this multi-bolt structure. Here, a method of calculating the bolt load ratio using four strain gauges is introduced. The positions of these four strain gauges on the composite laminates are shown in Fig.\ref{fig-stress}. To deeply expatiate on the feasibility of the method, a detailed stress analysis on the bolt hole is performed. For the external load applied to a composite joint, it can be seen as a combination of bearing load and by-pass load. Bearing load $P_{br}$ is the load transferred by the bolt while the by-pass load $P_{by}$ is the remaining load taken by the composite laminates. Due to the brittle nature of composites, a superposition principle can be used under linear assumption. According to Ref\cite{liu2013modified}, two non-dimensional stress concentration factors $\alpha_{br}$ and $\alpha_{by}$, can be utilized to describe the linear relation between the stress at a specific point and the applied external stress. Then, the stress at point 1 and point 2, can be written as,
\begin{equation}
\sigma_{1}=\alpha^{1}_{br}\cdot\sigma_{br}+\alpha^{1}_{by}\cdot\sigma_{by}
\label{E11}
\end{equation}
\begin{equation}
\sigma_{2}=-\alpha^{2}_{br}\cdot\sigma_{br}+\alpha^{2}_{by}\cdot\sigma_{by}
\label{E12}
\end{equation}
Here, for example, the $\alpha^{1}_{br}$ denotes for the stress concentration factor at point 1 under bearing. Bearing stress $\sigma_{br}=P_{br}/dt$ and by-pass stress $\sigma_{by}=P_{by}/(w-d)t$. As a result of the symmetric position of point 1 and point 2, $\alpha^{1}_{by}$ equals to $\alpha^{2}_{by}$. Therefore, bolt stress can be calculated from Eq.\ref{E11} and Eq.\ref{E12} as,
\begin{equation}
\sigma_{bolt1}=\sigma_{br}=\frac{\sigma_{1}-\sigma_{2}}{\alpha^{1}_{br}+\alpha^{2}_{br}}
\label{E13}
\end{equation}

For all the three bolts, similar equations can be obtained, thus making the bolt load ratio as,
\begin{equation}
ratio1=\frac{\sigma_{bolt1}}{\sigma_{bolt1}+\sigma_{bolt2}+\sigma_{bolt3}}=\frac{\sigma_{1}-\sigma_{2}}{\sigma_{1}-\sigma_{4}}
\label{E14}
\end{equation}
\begin{equation}
ratio2=\frac{\sigma_{bolt2}}{\sigma_{bolt1}+\sigma_{bolt21}+\sigma_{bolt3}}=\frac{\sigma_{2}-\sigma_{3}}{\sigma_{1}-\sigma_{4}}
\label{E15}
\end{equation}
\begin{equation}
ratio2=\frac{\sigma_{bolt3}}{\sigma_{bolt1}+\sigma_{bolt21}+\sigma_{bolt3}}=\frac{\sigma_{3}-\sigma_{4}}{\sigma_{1}-\sigma_{4}}
\label{E16}
\end{equation}

\begin{figure}[ht]
	\centering
	\includegraphics[width=13cm]{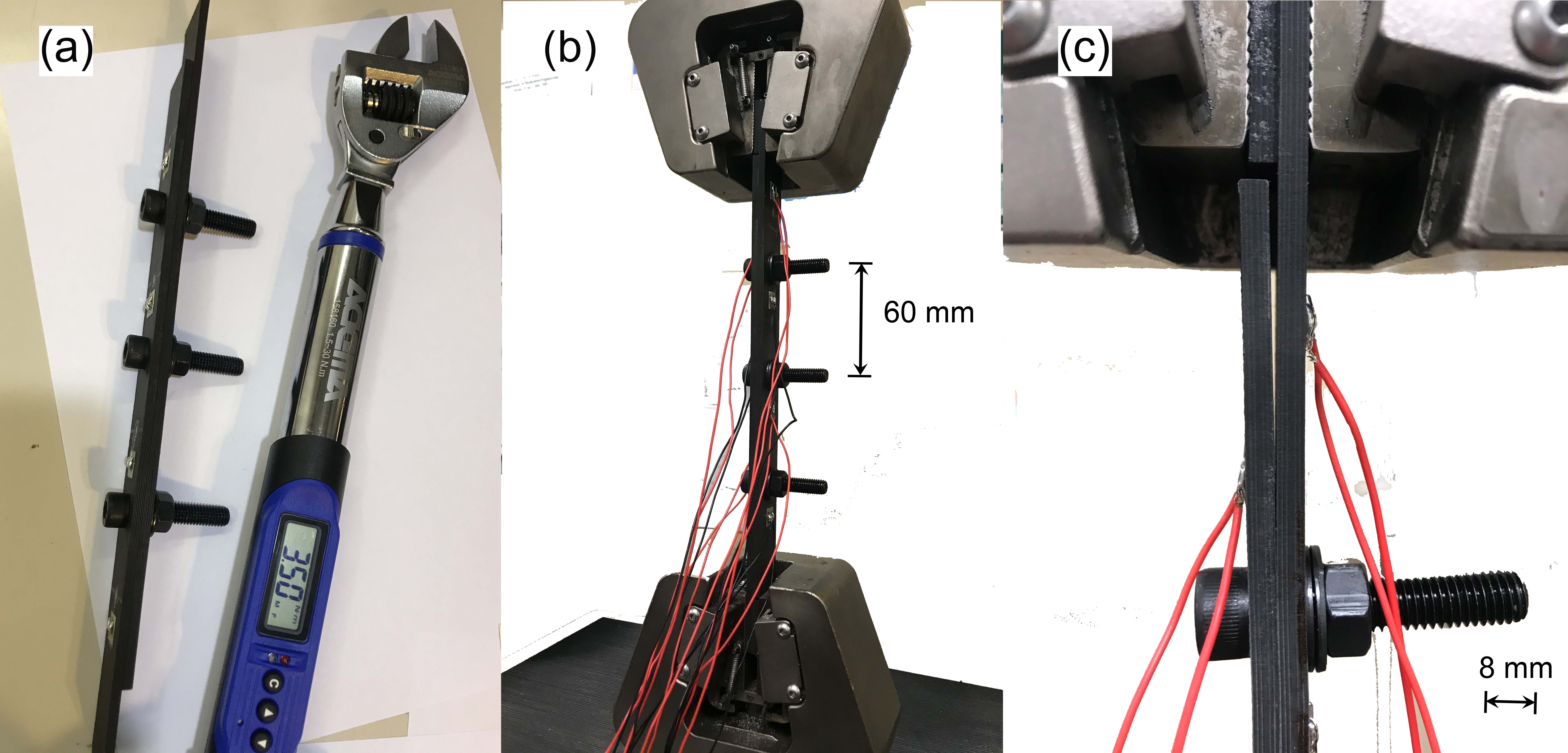}
	\caption{(a)The three-bolt joints cut by waterjet machine and assembled by a digital torque wrench (b)Experimental set-up (c)Secondary bending effect}
	\label{fig-setup}
\end{figure}

It can be seen from our method that the requirement for these four stain gauges is that they are placed symmetric to each other at the corresponding bolt hole. In our experiments, all the strain gauges are 30mm distant away from hole center. The bolt hole clearances are adjusted by a careful measurement of bolt diameter and the hole dimension cut by waterjet. The tightening torques of three bolts are ensured by a ETC Series digital torque wrench from ADEMA company. The fabricated three-bolt joints and the experimental set-up are shown in Fig.\ref{fig-setup}. In the experiments, a total external load of 30kN was applied and kept when the strain values at the four strain gauges at each joined plate were recorded and then calculated according to Eq.\ref{E14}-Eq.\ref{E16}.

\begin{figure}[ht]
	\centering
	\includegraphics[width=16cm]{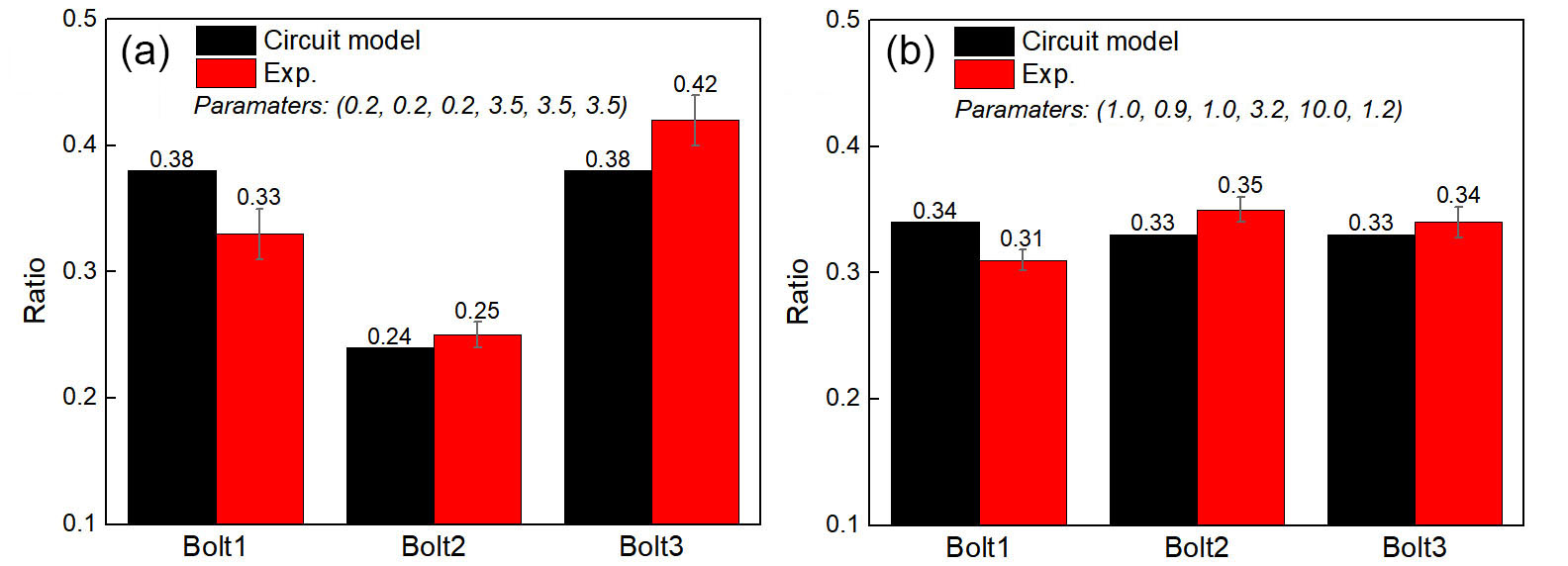}
	\caption{Experimental validation for the optimized design at an external tensile load of 30kN: comparison between results by circuit model and experiments under a three-bolt joint with (a)uniform parameters (b) parameters provided by optimization framework}
	\label{fig-exp}
\end{figure}

A total of four multi-bolt joints were made and follows two different parameters as (0.2mm, 0.2mm, 0.2mm, 3.5Nm, 3.5Nm, 3.5Nm) and (1.0mm, 0.9mm, 1.0mm, 3.2Nm, 10.0Nm, 1.2Nm). The experimental results calculated using Eq.\ref{E14}-Eq.\ref{E16} and results by circuit model are shown in Fig.\ref{fig-exp}. It can be shown that for these two configurations, results by circuit model agree well with the experimental method, especially at the middle bolt 2. For the bolt 1 and bolt 3, relatively large error is shown. This is due to the secondary bending effect as shown in Fig.\ref{fig-setup}. Normally, this bending would have effect on value of strain gauges 1 and 4. But at the strain gauge 2 and 3, this bending is restrained by the contact brought by tightening torques. Therefore, a better accuracy is shown in for the mid bolt. Comparing the results of these two configurations, it can be found that for the first group of specimens, bolt 1 and bolt 3 carry clearly more load than bolt 2. With parameters (1.0mm, 0.9mm, 1.0mm, 3.2Nm, 10.0Nm, 1.2Nm) provided by optimization method "GA+Circuit", results of the second group of specimens give a better even bolt-load-distribution as bolt 2 carries more load. This is achieved by relatively smaller the clearance and larger value of the tightening torque at bolt 2. However, due to the complex multi-stage pattern brought by friction effect, the mechanism of the combination of clearances and torques is not straightforward and this is where our machine learning model can play an important role.

\section{Conclusion}
 To conclude, in this paper, we have proposed an effective optimization method for the design of multi-bolt joint through a combination of a simple machine learning model and a novel equivalent circuit model. 
 
 The idea of this equivalent circuit model for determining load distribution in multi-bolt joint comes from the similarity between circuit system and linear mechanic system. According to this similarity, the load shared by each part of the joint structure is equal to the current flow in circuit model. Apart from the advantage that circuit model shows better computational efficiency, effect of frictions can also be reflected directly in this model by choosing various parameters indicating bolt-hole clearances and tightening-torques. Moreover, this circuit model is not limited to the single-bolt or three-bolt joint shown in the paper. It can be easily extended to joint with more bolts through serial and parallel modules.
 
 While the comparative study between identical and random parameters of clearances and torques has clearly shown that this friction effect has a great influence on the load distribution ratio on each bolt, we are focusing on reaching the minimum bolt load unevenness by carefully alter these two significant parameters of each bolt. 
 
 The machine-learning based framework here aims to provide both the forward response predictor and the backward problem of optimization. First, it is used to generate a symbol of load distribution unevenness from given set of parameters. Since that the dimensions of input and output is small, a very simple machine learning model using neural network with low computational cost is successfully established and validated. We then form a database which is composed of three bolt-hole clearances in the range (0,2mm), three tightening torques in the range (0,15Nm), and the corresponding symbol of unevenness of bolt-load-distribution. Then, with these database, for the optimization problem of load distribution unevenness, machine learning model could provide optimal parameters for achieving minimum unevenness. In addition, by a collaboration with optimization algorithms like GA and PSO, our results show that the computational efficiency is greatly increased, which is the most impressive advantages of our framework. 
 
 An experimental method of simply capturing the bolt load distribution using four strain gauges is introduced. This is  based on an assumption of linear relationship between bolt load and stress at a certain position. By adopting this experimental method, the output of our optimization framework is validated.
 
 The machine learning based framework can be a general procedure for accelerating the inverse problem elucidating structure-property relationship.  A possible application could be problems where a complicated relationship is hidden between input and output, or a vast design space is involved.

\section*{Data Availability}

The data that support the findings of this study are available from the corresponding author upon reasonable request.

\section*{Acknowledgement}

The authors are grateful to the support from NSFC/RGC Joint Research Scheme of Hong Kong (Grant \#: N\_HKUST 631/18), Nanhai-HKUST Program (Grant \#: FSNH-18FYTRI01).

\section*{Author contribution}
Jinglei Yang and Zhidong Guan proposed and designed the project. Cheng Qiu provided the idea of circuit model and performed benchmark calculation. Fengyang Jiang, Yuzi Han, Cheng Qiu developed machine-learning model. Shanyi du supervised the project. All authors contributed to the analysis of data as well as writing and revising of the manuscript.


\begin{thebibliography}{10}
	
	\bibitem{thoppul2009mechanics}
	Srinivasa~D Thoppul, Joana Finegan, and Ronald~F Gibson.
	\newblock Mechanics of mechanically fastened joints in polymer--matrix
	composite structures--a review.
	\newblock {\em Composites Science and Technology}, 69(3-4):301--329, 2009.
	
	\bibitem{katnam2013bonded}
	K~B Katnam, LFM Da~Silva, and TM~Young.
	\newblock Bonded repair of composite aircraft structures: A review of
	scientific challenges and opportunities.
	\newblock {\em Progress in Aerospace Sciences}, 61:26--42, 2013.
	
	\bibitem{raju2016improving}
	Karthik~Prasanna Raju, Kobye Bodjona, Gyu-Hyeong Lim, and Larry Lessard.
	\newblock Improving load sharing in hybrid bonded/bolted composite joints using
	an interference-fit bolt.
	\newblock {\em Composite Structures}, 149:329--338, 2016.
	
	\bibitem{camanho2006design}
	Pedro~Ponces Camanho and Michel Lambert.
	\newblock A design methodology for mechanically fastened joints in laminated
	composite materials.
	\newblock {\em Composites Science and Technology}, 66(15):3004--3020, 2006.
	
	\bibitem{xiao2005bearing}
	Yi~Xiao and Takashi Ishikawa.
	\newblock Bearing strength and failure behavior of bolted composite joints
	(part i: Experimental investigation).
	\newblock {\em Composites Science and Technology}, 65(7-8):1022--1031, 2005.
	
	\bibitem{xiao2005bearing2}
	Yi~Xiao and Takashi Ishikawa.
	\newblock Bearing strength and failure behavior of bolted composite joints
	(part ii: modeling and simulation).
	\newblock {\em Composites science and technology}, 65(7-8):1032--1043, 2005.
	
	\bibitem{puck2002failure}
	Alfred Puck and Helmut Sch{\"u}rmann.
	\newblock Failure analysis of frp laminates by means of physically based
	phenomenological models.
	\newblock {\em Composites science and technology}, 62(12-13):1633--1662, 2002.
	
	\bibitem{pinho2006physically}
	ST~Pinho, L~Iannucci, and P~Robinson.
	\newblock Physically based failure models and criteria for laminated
	fibre-reinforced composites with emphasis on fibre kinking. part ii: Fe
	implementation.
	\newblock {\em Composites Part A: Applied Science and Manufacturing},
	37(5):766--777, 2006.
	
	\bibitem{mccarthy2006simple}
	MA~McCarthy, CT~McCarthy, and GS~Padhi.
	\newblock A simple method for determining the effects of bolt--hole clearance
	on load distribution in single-column multi-bolt composite joints.
	\newblock {\em Composite Structures}, 73(1):78--87, 2006.
	
	\bibitem{lecomte2014analytical}
	J~Lecomte, C~Bois, H~Wargnier, and J-C Wahl.
	\newblock An analytical model for the prediction of load distribution in
	multi-bolt composite joints including hole-location errors.
	\newblock {\em Composite Structures}, 117:354--361, 2014.
	
	\bibitem{yang2018enhanced}
	Yuxing Yang, Xueshu Liu, Yi-Qi Wang, Hang Gao, Yongjie Bao, and Rupeng Li.
	\newblock An enhanced spring-mass model for stiffness prediction in single-lap
	composite joints with considering assembly gap and gap shimming.
	\newblock {\em Composite Structures}, 187:18--26, 2018.
	
	\bibitem{liu2018interpretation}
	Fengrui Liu, Xuheng Lu, Libin Zhao, Jianyu Zhang, Ning Hu, and Jifeng Xu.
	\newblock An interpretation of the load distributions in highly torqued
	single-lap composite bolted joints with bolt-hole clearances.
	\newblock {\em Composites Part B: Engineering}, 138:194--205, 2018.
	
	\bibitem{zhao2019modified}
	Libin Zhao, Ziang Fang, Fengrui Liu, Meijuan Shan, and Jianyu Zhang.
	\newblock A modified stiffness method considering effects of hole tensile
	deformation on bolt load distribution in multi-bolt composite joints.
	\newblock {\em Composites Part B: Engineering}, 171:264--271, 2019.
	
	\bibitem{mccarthy2011analytical}
	CT~McCarthy and PJ~Gray.
	\newblock An analytical model for the prediction of load distribution in highly
	torqued multi-bolt composite joints.
	\newblock {\em Composite Structures}, 93(2):287--298, 2011.
	
	\bibitem{qiu2019improved}
	Cheng Qiu, Zhidong Guan, Shanyi Du, and Zengshan Li.
	\newblock An improved model for predicting stiffness of single-lap composites
	bolted joints using matlab/simulink.
	\newblock {\em Mechanics of Advanced Materials and Structures}, pages 1--11,
	2019.
	
	\bibitem{mccarthy2005three}
	CT~McCarthy and MA~McCarthy.
	\newblock Three-dimensional finite element analysis of single-bolt, single-lap
	composite bolted joints: Part ii----effects of bolt-hole clearance.
	\newblock {\em Composite Structures}, 71(2):159--175, 2005.
	
	\bibitem{schmidt2019recent}
	Jonathan Schmidt, M{\'a}rio~RG Marques, Silvana Botti, and Miguel~AL Marques.
	\newblock Recent advances and applications of machine learning in solid-state
	materials science.
	\newblock {\em npj Computational Materials}, 5(1):1--36, 2019.
	
	\bibitem{chen2019machine}
	Chun-Teh Chen and Grace~X Gu.
	\newblock Machine learning for composite materials.
	\newblock {\em MRS Communications}, 9(2):556--566, 2019.
	
	\bibitem{ramprasad2017machine}
	Rampi Ramprasad, Rohit Batra, Ghanshyam Pilania, Arun Mannodi-Kanakkithodi, and
	Chiho Kim.
	\newblock Machine learning in materials informatics: recent applications and
	prospects.
	\newblock {\em npj Computational Materials}, 3(1):1--13, 2017.
	
	\bibitem{zobeiry2020theory}
	Navid Zobeiry, Johannes Reiner, and Reza Vaziri.
	\newblock Theory-guided machine learning for damage characterization of
	composites.
	\newblock {\em Composite Structures}, page 112407, 2020.
	
	\bibitem{ward2016general}
	Logan Ward, Ankit Agrawal, Alok Choudhary, and Christopher Wolverton.
	\newblock A general-purpose machine learning framework for predicting
	properties of inorganic materials.
	\newblock {\em npj Computational Materials}, 2(1):1--7, 2016.
	
	\bibitem{huber2018machine}
	Liam Huber, Raheleh Hadian, Blazej Grabowski, and J{\"o}rg Neugebauer.
	\newblock A machine learning approach to model solute grain boundary
	segregation.
	\newblock {\em npj Computational Materials}, 4(1):1--8, 2018.
	
	\bibitem{yang2018deep}
	Zijiang Yang, Yuksel~C Yabansu, Reda Al-Bahrani, Wei-keng Liao, Alok~N
	Choudhary, Surya~R Kalidindi, and Ankit Agrawal.
	\newblock Deep learning approaches for mining structure-property linkages in
	high contrast composites from simulation datasets.
	\newblock {\em Computational Materials Science}, 151:278--287, 2018.
	
	\bibitem{ma2020accelerated}
	Chunping Ma, Zhiwei Zhang, Benjamin Luce, Simon Pusateri, Binglin Xie,
	Mohammad~H Rafiei, and Nan Hu.
	\newblock Accelerated design and characterization of non-uniform cellular
	materials via a machine-learning based framework.
	\newblock {\em npj Computational Materials}, 6(1):1--8, 2020.
	
	\bibitem{liu2015predictive}
	Ruoqian Liu, Abhishek Kumar, Zhengzhang Chen, Ankit Agrawal, Veera
	Sundararaghavan, and Alok Choudhary.
	\newblock A predictive machine learning approach for microstructure
	optimization and materials design.
	\newblock {\em Scientific reports}, 5(1):1--12, 2015.
	
	\bibitem{chen2020generative}
	Chun-Teh Chen and Grace~X Gu.
	\newblock Generative deep neural networks for inverse materials design using
	backpropagation and active learning.
	\newblock {\em Advanced Science}, 7(5):1902607, 2020.
	
	\bibitem{gu2018bioinspired}
	Grace~X Gu, Chun-Teh Chen, Deon~J Richmond, and Markus~J Buehler.
	\newblock Bioinspired hierarchical composite design using machine learning:
	simulation, additive manufacturing, and experiment.
	\newblock {\em Materials Horizons}, 5(5):939--945, 2018.
	
	\bibitem{ramasamy2014prediction}
	P~Ramasamy and S~Sampathkumar.
	\newblock Prediction of impact damage tolerance of drop impacted wgfrp
	composite by artificial neural network using acoustic emission parameters.
	\newblock {\em Composites Part B: Engineering}, 60:457--462, 2014.
	
	\bibitem{tran2020active}
	Anh Tran, John~A Mitchell, Laura Swiler, and Tim Wildey.
	\newblock An active learning high-throughput microstructure calibration
	framework for solving inverse structure-process problems in materials
	informatics.
	\newblock {\em Acta Materialia}, 2020.
	
	\bibitem{mccarthy2005experiences}
	CT~McCarthy, MA~McCarthy, WF~Stanley, and VP~Lawlor.
	\newblock Experiences with modeling friction in composite bolted joints.
	\newblock {\em Journal of composite materials}, 39(21):1881--1908, 2005.
	
	\bibitem{abaqus2007abaqus}
	Inc Abaqus.
	\newblock Abaqus user’s manual.
	\newblock {\em ABAQUS Ine, Dassault Systemes, Rhode Island, USA}, 2007.
	
	\bibitem{liu2013modified}
	Fengrui Liu, Libin Zhao, Saqib Mehmood, Jianyu Zhang, and Binjun Fei.
	\newblock A modified failure envelope method for failure prediction of
	multi-bolt composite joints.
	\newblock {\em Composites Science and Technology}, 83:54--63, 2013.
	
\end{thebibliography}
\end{document}